%% file: code2math_iclr2027.tex
\documentclass{article}
\usepackage{iclr2027_conference,times}
\iclrfinalcopy
\input{math_commands.tex}

\usepackage[T1]{fontenc}
\usepackage[utf8]{inputenc}
\usepackage{hyperref}
\usepackage{graphicx}
\usepackage{caption}
\usepackage{microtype}
\usepackage{booktabs}
\usepackage{amsmath}
\usepackage{textcomp}
\usepackage{amssymb}
\usepackage{mathtools}
\usepackage{amsthm}
\usepackage{multirow}
\usepackage{wrapfig}
\usepackage[most]{tcolorbox}
\newtcolorbox{showcase}[2][]{colback=gray!5!white,
  colframe=blue!50!black,
  title={#2},
  fonttitle=\bfseries,
  sharp corners,
  boxrule=1pt,
  left=2mm, right=2mm, top=1mm, bottom=1mm,
  enhanced,
  breakable,
  #1
}
\title{Code2Math: Can Your Code Agent Evolve Math Problems Through Exploration?}

\author{
 \textbf{Dadi Guo\textsuperscript{1,*}},
 \textbf{Yuejin Xie\textsuperscript{2,*}},
 \textbf{Qingyu Liu\textsuperscript{3,*}},
 \textbf{Weixian Huang\textsuperscript{4,*}},
 \textbf{Jiayu Liu\textsuperscript{1}},
 \textbf{Zhiyuan Fan\textsuperscript{1}},
\\
 \textbf{Qihan Ren\textsuperscript{5}},
 \textbf{Shuai Shao \textsuperscript{5}},
 \textbf{Tianyi Zhou\textsuperscript{6}},
 \textbf{Jianjie Feng\textsuperscript{7}},
 \textbf{Wenze Su\textsuperscript{2}},
\\
 \textbf{Yujiu Yang\textsuperscript{2}},
 \textbf{Dongrui Liu\textsuperscript{5}},
 \textbf{Yi R.(May) Fung\textsuperscript{1}}
\\
\\
 \textsuperscript{1}Hong Kong University of Science and Technology,
 \textsuperscript{2}Tsinghua University,
 \textsuperscript{3}Zhejiang University,
\\
 \textsuperscript{4}Nanjing Tech University,
 \textsuperscript{5}Shanghai Jiao Tong University,
 \textsuperscript{6}University of Michigan,
\\
 \textsuperscript{7}Independent Researcher
\\
 \small{*Equal contribution, 
   Correspondence: {yrfung@ust.hk, drliu96@sjtu.edu.cn}
 }
}

\begin{document}
\maketitle
\begin{abstract}
While large language models have achieved remarkable progress in solving mathematical problems, reaching IMO-level performance and tackling research-level tasks, the ability to formulate new and more challenging mathematical problems remains an important yet underexplored capability.
This capability is particularly relevant to recursive self-improvement, where models are expected to autonomously expand the scope and difficulty of the tasks they can engage with.
Meanwhile, recent code agents provide executable environments for symbolic computation, systematic search, and empirical verification, making them a natural substrate for autonomous mathematical exploration.
To this end, we introduce Code2Math, a multi-agent system that leverages code agents to evolve mathematical problems through executable test-time exploration.
Our experiments demonstrate that, given sufficient test-time exploration, code agents can synthesize new, solvable problems that are structurally distinct from and more challenging than the originals.
Moreover, augmenting the original training data with the corresponding evolved versions of these problems further improves model performance compared with training on the original data alone, demonstrating the practical value of problem evolution for model training.
Together, these results demonstrate the potential of code-driven problem evolution as a scalable mechanism for autonomous mathematical exploration and a building block toward self-evolving reasoning systems. Code and data is available at \url{https://github.com/TarferSoul/Code2Math}.
\end{abstract}

\section{Introduction}
Recent large language models (LLMs) and agents have achieved substantial advances in mathematical reasoning, reaching IMO-level performance and increasingly tackling research-level mathematical problems \citep{openai2026navierstokes,huang2025winning, deepseek-math-v2, DeepSeekAI2025DeepSeekV32PT, Gao2025LonghorizonRA}. 
Despite this rapid progress in problem solving, comparatively less attention has been paid to the complementary ability of models to formulate new mathematical problems. 
Problem formulation requires identifying meaningful structures, modifying constraints, and anticipating how these changes alter the reasoning process, making it a distinct capability from solving a fixed instance. 
Developing such capabilities is also closely connected to recursive self-improvement (RSI)~\citep{Meng2026RSIBenchDataBD, Yang2026DataFoundryED}, where a system should be able to create increasingly informative challenges rather than rely solely on a static collection of human-curated tasks. 
This motivates automated methods for mathematical problem evolution that can continuously produce new, nontrivial problems for both evaluation and learning.

\begin{wrapfigure}{r}{0.52\columnwidth}
    \centering
    \vspace{-6pt}
    \includegraphics[width=\linewidth]{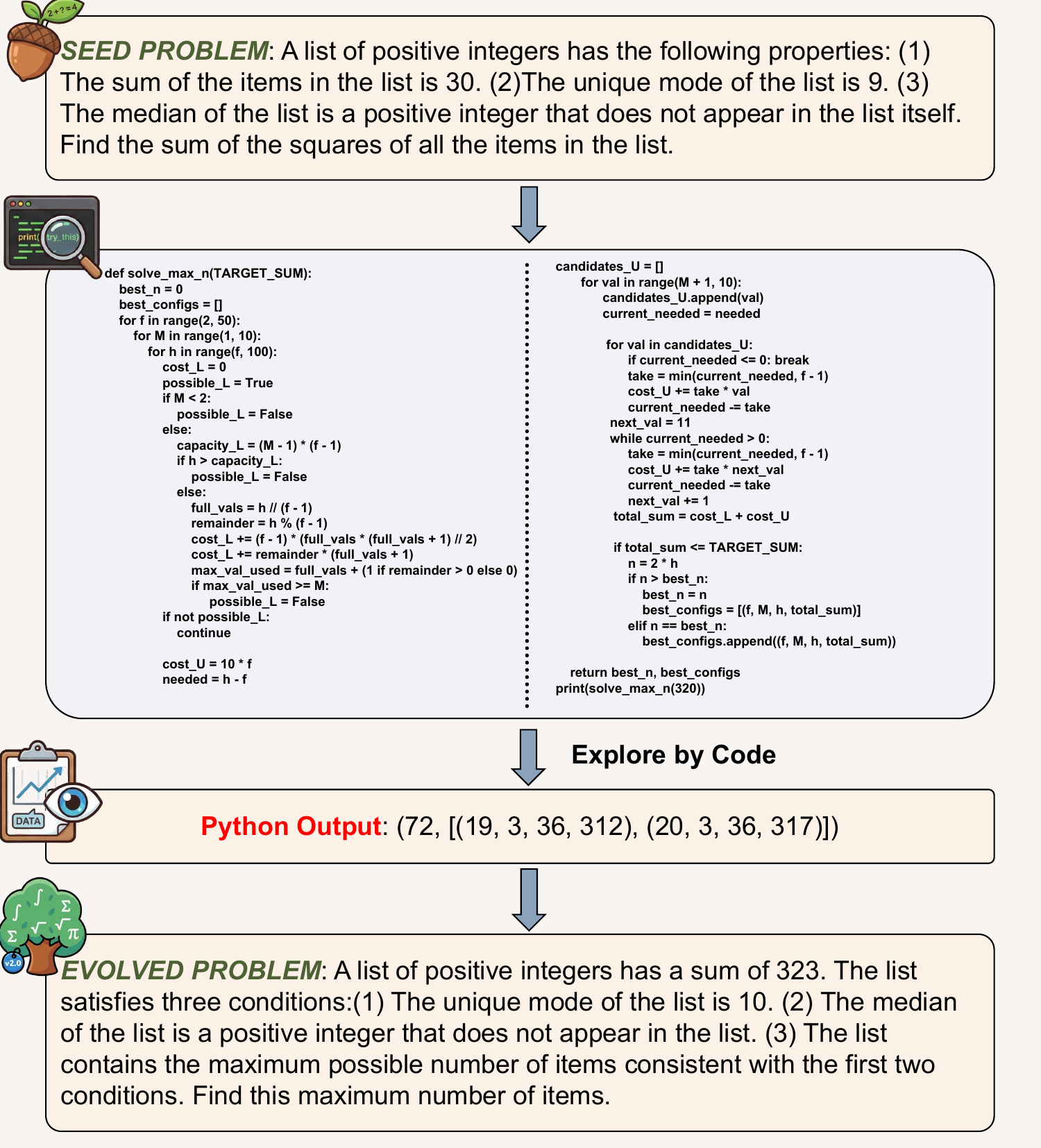}
    \caption{Example of code-driven problem evolution. 
    }
    \label{fig:first_fig}
    \vspace{-10pt}
\end{wrapfigure}

Formulating challenging mathematical problems often involves exploratory processes such as conjecture formation, counterexample search, and systematic experimentation over structured spaces. 
These processes are inherently computational, involving iterative hypothesis testing and verification rather than purely deductive reasoning. 
Recent advances in code agents \citep{yang2025codefoundationmodelsagents, DeepSeekAI2025DeepSeekV32PT} combine strong reasoning capabilities with access to scalable and executable computational environments \citep{huang2025environment}, enabling large-scale simulation, symbolic manipulation, and automated verification \citep{cheng2026llminsandboxelicitsgeneralagentic}. 
As illustrated in Figure~\ref{fig:first_fig}, a code agent can systematically enumerate valid configurations under structural constraints, identify extremal patterns from the resulting observations, and then abstract these findings into a new mathematical problem.
Such capabilities closely mirror the workflows through which human mathematicians discover and refine new problems. 
This alignment suggests that code agents provide a promising mechanism for autonomously exploring mathematical spaces and synthesizing novel, challenging problems, offering a scalable source of high-quality mathematical reasoning data.

In this paper, we investigate whether code agents can autonomously evolve existing mathematical problems into new, more challenging ones. We aim to answer three research questions: (1) Can evolved problems remain mathematically valid while becoming genuinely more challenging? (2) How efficiently can code agents produce such problems? (3) Do the evolved problems provide useful additional training signals beyond the original data?

We propose \textsc{Code2Math}, a multi-agent framework for evolving mathematical problems through code-driven exploration and verification. 
The framework consists of three specialized agents: the \textit{Evolution Agent}, the \textit{Solvability Verification Agent}, and the \textit{Difficulty Verification Agent}. 
Inspired by Theory of Mind \citep{chen2025theory,tell-me-more,userbench,marcon}, the \textit{Evolution Agent} anticipates likely solver strategies and explores modifications that introduce new \textit{Aha moments} \citep{guo2025deepseek}, making the key insight more difficult to discover. 
The \textit{Solvability Verification Agent} checks the evolved problem and its proposed solution for consistency and logical validity, rejecting candidates with flawed solution paths. 
Finally, the \textit{Difficulty Verification Agent} assesses whether the evolved problem increases the \textit{Burden of Discovery}, which we define as the difficulty of identifying the key insight required for solving the problem. Our system follows the test-time scaling paradigm \citep{muennighoff2025s1, zhang2025survey} by using multiple rollouts \cite{multirole-r1} from the \textit{Evolution Agent} until both verification agents' criteria are satisfied. This gives the agent sufficient room for exploration while also yielding efficiency metrics for problem creation, such as rollout count. Unlike scaling methods based only on text-based long-chain reasoning \citep{wei2022chain}, our agent can write code and use mathematical Python libraries such as \textit{SymPy}, \textit{NetworkX}, and \textit{itertools} \citep{li2023chain}, enabling symbolic computation and deterministic intermediate feedback to guide evolution.

We evaluate \textsc{Code2Math} across five evolution models and six solver models. 
Our experiments show that the evolved problems remain highly solvable while introducing substantial increases in difficulty across different solvers. 
We further find that successful evolution typically requires multiple exploration attempts, revealing a nontrivial computational cost of generating qualified problems. 
Finally, augmenting the original training data with evolved problems consistently improves model performance across three AIME-series benchmarks, demonstrating that the generated problems provide useful training signals beyond their role as challenging evaluation instances.
Our contributions are as follows:

\begin{itemize}

\item We propose \textsc{Code2Math}, a multi-agent framework that leverages executable exploration to evolve mathematical problems, with dedicated agents for evolution, solvability verification, and difficulty verification.

\item We conduct extensive evaluations across multiple evolution and solver models, showing that Code2Math generates mathematically valid problems with substantially increased difficulty, while characterizing the exploration cost required for successful evolution.

\item We further demonstrate the downstream utility of evolved problems: augmenting the original training data with evolved problems consistently improves model performance across three AIME-series benchmarks.

\end{itemize}
\section{Related Works}
\textbf{Data Synthesis through Exploration.}
Recent studies have leveraged models' environment exploration ability to synthesize new data. AgentEvolver \citep{AgentEvolver2025}, WebExplorer \citep{Liu2025WebExplorerEA}, TaskCraft \citep{Shi2025TaskCraftAG}, Go-Browse \citep{gandhi2025go}, and Cognitive Kernel-Pro \citep{fang2025cognitive} enable models to explore environments and progressively generate agent data. TRACE \citep{Guo2025TowardsSB} and AutoCode \citep{zhou2025autocode} further show that multi-agent systems can evolve general-purpose and coding tasks through exploration with verification. However, these works mainly focus on agent task generation and rarely address mathematical reasoning tasks. AlphaGeometry \citep{trinh2024solving} explores new geometric problems from known structures, but relies on a specialized symbolic deduction engine for geometry.

\noindent\textbf{Math Problem Adaptation and Generation.}
Prior work has adapted mathematical problems into new training or benchmark data. MATH-Perturb \citep{huang2025math}, EvolMathEval \citep{wang2025evolmatheval}, and Benchmark Self-Evolving \citep{wang2025benchmark} modify existing benchmark problems to evaluate reasoning robustness. However, they often depend on manual effort or simple rule-based LLM edits, leaving the agentic potential of models underexplored. Another line of work, including R-zero \citep{huang2025r}, Self-Question Language Model \citep{chen2025self}, UltraLogic \citep{liu2026ultralogic}, SANDMath \citep{manem2025sand}, and RLVE \citep{zeng2025rlve}, directly generates math problems for training. While effective, these methods make limited use of agentic exploration and may lack systematic evaluation of generated problem quality.

\section{Method}
In this section, we introduce the seed problems used for evolution, the multi-agent framework including three different agents, and the evaluation method.

\subsection{Data for Evolution}
\label{sec:data}
We use two separate data collections for problem evolution and downstream training evaluation. For evaluating the problem-evolution capability of Code2Math, we collect 100 seed problems from diverse sources and difficulty levels, covering algebra, combinatorics, calculus, sequences, graph theory, and other domains. The problems are drawn from standard mathematics problem books, regional examinations and competitions, the IMO, and common benchmarks such as AIME-2024 and AIME-2025. This diverse seed set is designed to test whether the framework can reliably evolve problems across different mathematical domains and levels of difficulty. We additionally select six problem pairs as demonstrations for problem adaptation. These include expert-written examples as well as pairs generated through reverse creation \citep{sun2025genesis}, where a simpler problem is first derived from a more complex one and the corresponding adaptation process is then provided as a demonstration. For the downstream training experiment, we separately collect 2,500 problems from publicly available mathematical datasets~\citep{Ren2026RethinkingGI}. These problems are used to construct the original training set and the corresponding evolved training set, allowing us to evaluate whether augmenting the original data with Code2Math-evolved problems improves model performance.

\subsection{Multi-Agent System}
\begin{figure}[t]
  \centering
  \includegraphics[width=0.9\linewidth]{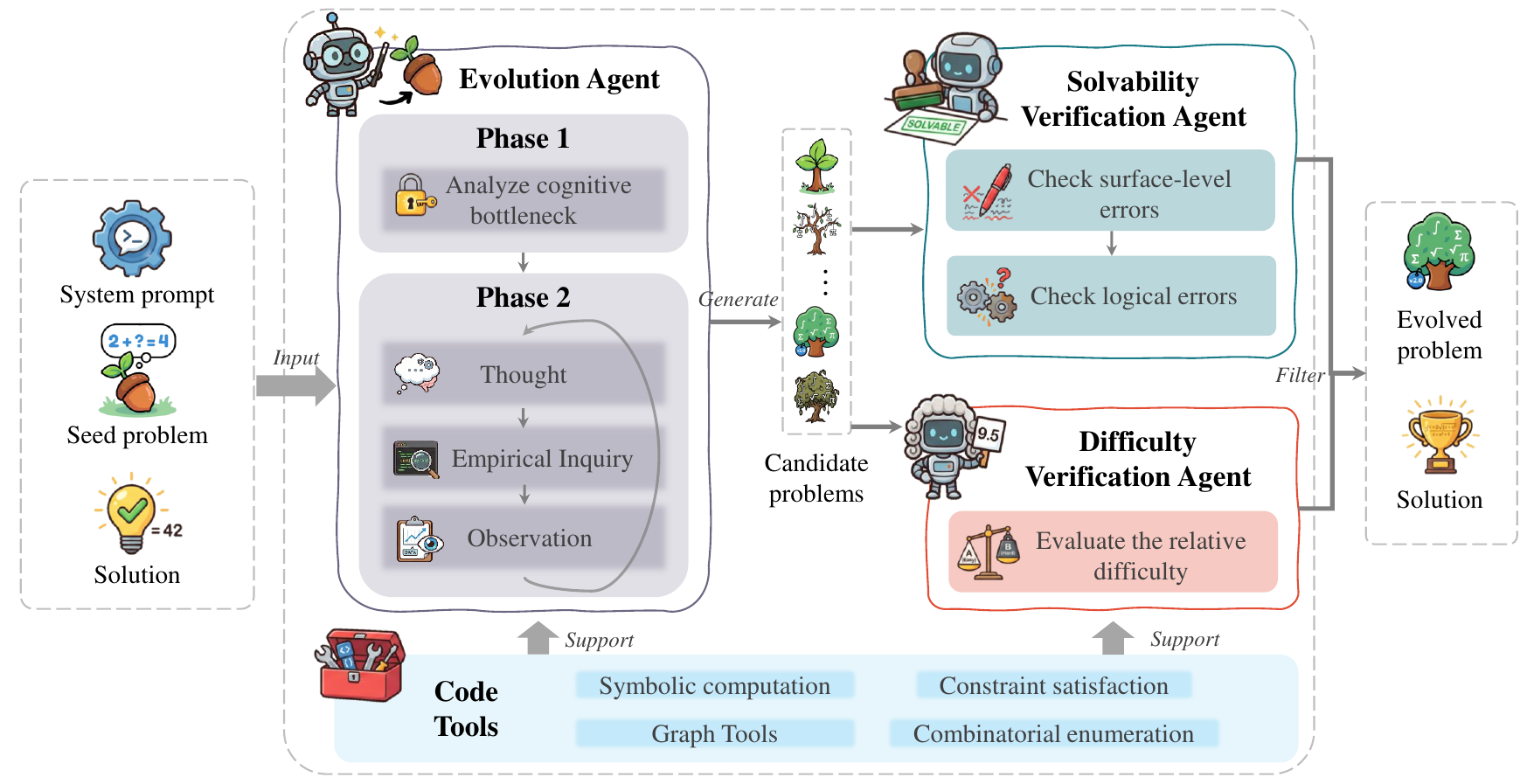}
  \caption{Overview of our multi-agent system. Our pipeline consists of three components: the \textit{Evolution Agent}, the \textit{Solvability Verification Agent}, and the \textit{Difficulty Verification Agent}. It is equipped with code tools related to mathematics. 
  The framework takes an original problem and its solution as input and outputs a validated new problem along with a solution for reference.}
  \label{fig:overview}
\end{figure}

\textsc{Code2Math} consists of three agents: the \textit{Evolution Agent}, the \textit{Solvability Verification Agent}, and the \textit{Difficulty Verification Agent}. In this section, we introduce the design of each agent individually. The overview of \textsc{Code2Math} is shown in Figure \ref{fig:overview}.

\textbf{Evolution Agent.} The \textit{Evolution Agent} takes a seed problem and its solution steps as input. We design this agent to operate in two phases. In the first phase, it analyzes the solution of the input problem to identify the cognitive bottleneck for a solver. In the second phase, it performs free exploration based on the original problem to design a more challenging new problem. Drawing inspiration from Theory of Mind \citep{chen2025theory}, we define difficulty here as the \textit{Burden of Discovery}. Specifically, we require the agent to anticipate how an experienced competition solver would approach the problem, and then deliberately conceal potential insights to make them difficult to uncover, thereby creating an \textit{Aha moment} \citep{guo2025deepseek} within the solution process of the new problem. Additionally, we offer guidance on potential directions by encouraging the agent to explore areas such as tighter mathematical bounds, more sophisticated combinatorial constructions, and underlying patterns in numerical sequences. The output of the \textit{Evolution Agent} comprises the new problem statement and the proposed solution steps.

\textbf{Solvability Verification Agent.}
Determining problem solvability \citep{peng2025learning,naacl} is non-trivial, as flaws may include not only surface-level errors in the problem statement but also subtle logical inconsistencies. We therefore design the \textit{Solvability Verification Agent} to operate in two stages: first detecting obvious surface-level errors, then scrutinizing the solution steps proposed by the \textit{Evolution Agent}. The rationale is that a flawless logical chain implies the existence of at least one solution path, serving as a proxy for solvability. Conversely, if the proposed solution contains logical flaws, we discard the problem even if it might be intrinsically solvable. Since many selected problems lack deterministic final answers (\textit{e.g.}, proof problems), we use a set of failure modes \citep{guo2025mathematical, Yuan2025CuringMS,naacl} to help the agent diagnose and categorize solution errors.

\textbf{Difficulty Verification Agent.}
The \textit{Difficulty Verification Agent} receives the original and adapted problems, along with their respective solution steps. Following the ToM-based approach above, it evaluates whether the adapted problem introduces an \textit{Aha moment} that is harder for an experienced competition solver to uncover.

To rigorously assess adaptation quality, the agent uses a 5-point scoring mechanism that distinguishes \textit{Artificial Complexity} from \textit{Cognitive Depth}. Scores 1--2 fall below the acceptance threshold and indicate failures to induce genuine difficulty: Score 1 corresponds to an unchanged or regressed solution path, while Score 2 captures difficulty increases caused merely by computational tedium or repetitive procedures. Penalizing such cases prevents the agent from mistaking labor-intensive algebra for intellectual challenge.

Successful adaptations start at Score 3, where the problem breaks standard solution templates and forces a deviation from rote application. Scores 4--5 are reserved for stronger anti-templating adaptations that turn standard heuristics into traps and require deeper \textit{Aha moments}. The highest tier, Score 5, further rewards mathematical beauty, such as deep symmetries or unexpected conceptual connections.

\subsection{Test-time Exploration through Code}
Leveraging the test-time scaling paradigm, our approach involves generating multiple rollouts per input to satisfy the criteria of our dual-verification system. Agents are explicitly governed to utilize code as a tool for empirical inquiry. For instance, they may run numerical simulations to probe for tighter inequality bounds, print out sequences to intuitively spot regularities, or validate their hypotheses by actively searching for counter-examples. We equip the agent with a comprehensive Python sandbox containing a curated suite of libraries spanning symbolic computation (\textit{SymPy}), constraint satisfaction (\textit{Z3}), graph theory (\textit{NetworkX}), and combinatorial enumeration (\textit{itertools}). This rich toolset empowers the agent to perform rigorous empirical verification across diverse mathematical domains, ranging from high-precision arithmetic to complex topological analysis.

\subsection{Evaluation Method}
We conduct evaluations on the solvability of the generated questions, the increase in difficulty, the efficiency of the model in evolving questions, and the role of code during exploration, respectively.

\textbf{Solvability.} There is no deterministic way to assess the solvability of a natural language math problem, so we adopt an LLM-as-a-judge approach \citep{gu2024survey}. Our earlier multi-agent framework already includes a solvability check, but frameworks based on different models can vary a lot. Therefore, we introduce a unified third-party model to perform the check in a consistent manner. To maximize the reliability of the evaluation, this third party must be a model with strong and dependable reasoning ability (in our experiments, we chose \textit{GPT-5.2-high}) \citep{oai_5_2_system-card}. A problem is deemed solvable only when the third-party judge and the agent reach agreement.

\textbf{Difficulty.} We evalaute different models on the original questions and on the new questions that pass the solvability check, observing whether their accuracy and reasoning length change. Lower accuracy and more reasoning tokens indicate that the evolved questions are harder. Accuracy is also evaluated using \textit{GPT-5.2-high}  as the judge to determine whether there are logical errors in the solution process.

\textbf{Efficiency.}
We use the average number of \textit{Evolution Agent} rollouts to get a qualified new question as a metric to evaluate the agent’s efficiency. In addition, we also compile the distribution of rollout counts across different models as an auxiliary metric.

\section{Experiments}
We first introduce the experimental setup, followed by results corresponding to our three research questions. We evaluate the validity and difficulty of evolved problems on the 100-problem benchmark, examine their downstream training utility using a separate 2,500-problem training corpus, and finally analyze the efficiency of the problem-evolution process. 

\subsection{Setup}

\textbf{Models.} We use five backbones for the evolution pipeline: \textit{DeepSeek-Chat}, \textit{DeepSeek-Reasoner}~\citep{DeepSeekAI2025DeepSeekV32PT}(note that the two models are DeepSeek-v3.2 version), \textit{Gemini-3-Pro-Preview-Thinking}~\citep{gemini3pro_modelcard}, \textit{Kimi-K2-Thinking}~\citep{team2025kimi}, and \textit{Seed-2.0-Pro}~\citep{seed2_model_card}. We evaluate the resulting problems with six solver models: \textit{DeepSeek-Chat}, \textit{DeepSeek-Reasoner}, \textit{Qwen3-235B-A22B-Thinking-2507}~\citep{Yang2025Qwen3TR}, \textit{Gemini-3-Flash-Thinking}, \textit{GPT-5.2-Medium}, and \textit{GPT-5.2-High}~\citep{oai_5_2_system-card}. We use \textit{GPT-5.2-High} as the external judge for solvability and answer correctness; when it also appears as a solver, judge calls are run as separate invocations.
For the downstream training experiments, we use \textit{GLM-5.3-Flash} ~\citep{glm5team2026glm5vibecodingagentic} to evolve the 2,500-problem training corpus, and train \textit{Qwen3-8B} and \textit{Qwen3-14B} under supervised fine-tuning~\citep{Yang2025Qwen3TR}. For both the original and evolved problems, the chain-of-thought solutions used as SFT targets are generated by \textit{Qwen3-30B}. During supervised fine-tuning, we train for 6 epochs with a learning rate of $5\times10^{-5}$ and a batch size of 32.

\textbf{Agentic Environment.} Our multi-agent system is implemented with Smolagents~\citep{smolagents}, where agents can execute Python code in a controlled environment. The sandbox provides standard Python utilities, scientific and symbolic computation packages, constraint solving, and graph-analysis tools; the full package list is provided in Appendix~\ref{app:setup-details}.

\textbf{Implementation Details.} We use 100 seed problems, with each seed assigned one evolution run. Each run allows at most 20 rollouts and a maximum trajectory length of 30 steps; a run succeeds only if a generated problem passes both verification agents. During solver evaluation, each model receives up to three attempts per problem and is stopped after either reaching its maximum token limit or a 30-minute wall-clock timeout. All models are queried with temperature 0.

\textbf{Evaluation Metrics.} We report Evolution Success Count (\textbf{ESC}), Certified Solvability Count (\textbf{CSC}), Agreement Rate (\textbf{AR}) between internal and external solvability checks, Solve Rate (\textbf{SR}) on both seed and evolved problems, and Average Token Consumption (\textbf{ATC}). Lower Evolution-SR than Origin-SR and higher token consumption indicate increased difficulty. For downstream training evaluation, we report Avg@10 accuracy on AIME-2024, AIME-2025, and AIME-2026.

\begin{table}[t]
\caption{Cross-model evaluation of problem evolution effectiveness. Column groups indicate the LLM used in the evolutionary phase, while rows indicate the solver model used in the evaluation phase. Each \textbf{AR} cell lists the number of evolved problems certified by the external judge, followed by the number that passed internal verification. \textbf{Origin-SR}(\%) and \textbf{Evolution-SR}(\%) represent the solve rates on the original seed problems and the evolved problems, respectively. The difference (\textit{Evolution-SR $-$ Origin-SR}) measures the extent of problem evolution, where a smaller (more negative) value indicates a greater increase in difficulty introduced by the evolutionary process.}
\vspace{-2mm}
\small
\centering
\setlength\tabcolsep{1.5pt}
\resizebox{\linewidth}{!}{
\begin{tabular}{llccccccccccccccc}
\toprule
\multicolumn{2}{l}{\multirow{2}{*}{Solver model in evaluation phase}} & \multicolumn{3}{c}{DeepSeek-Chat} & \multicolumn{3}{c}{DeepSeek-Reasoner} & \multicolumn{3}{c}{Gemini-3-Pro-Thinking} & \multicolumn{3}{c}{Kimi-K2-Thinking} & \multicolumn{3}{c}{Seed-2.0-Pro} \\\cmidrule(r){3-5}\cmidrule(r){6-8}\cmidrule(r){9-11}\cmidrule(r){12-14}\cmidrule(r){15-17}
\multicolumn{2}{l}{}                                               & AR     & Origin-SR & Evolution-SR & AR      & Origin-SR & Evolution-SR & AR      & Origin-SR & Evolution-SR & AR      & Origin-SR & Evolution-SR & AR      & Origin-SR & Evolution-SR \\\midrule
\multirow{3}{*}{\begin{tabular}[c]{@{}l@{}}Open-\\source\end{tabular}}   
& DeepSeek-Chat             & 83/94  & 16        & 19{\tiny \textcolor{gray}{$+3$}}  & 94/98   & 16        & 9{\tiny \textcolor{red}{$-7$}}    & 98/98   & 16        & 7{\tiny \textcolor{red}{$-9$}}    & 74/90   & 16        & 11{\tiny \textcolor{red}{$-5$}}   & 83/97   & 16        & 9{\tiny \textcolor{red}{$-7$}}    \\
& DeepSeek-Reasoner         & 83/94  & 48        & 34{\tiny \textcolor{red}{$-14$}}  & 94/98   & 48        & 36{\tiny \textcolor{red}{$-12$}}  & 98/98   & 48        & 8{\tiny \textcolor{red}{$-40$}}   & 74/90   & 48        & 39{\tiny \textcolor{red}{$-9$}}   & 83/97   & 48        & 26{\tiny \textcolor{red}{$-22$}}  \\
& Qwen3-235B-A22B-Thinking  & 83/94  & 20        & 15{\tiny \textcolor{red}{$-5$}}   & 94/98   & 20        & 19{\tiny \textcolor{red}{$-1$}}   & 98/98   & 20        & 18{\tiny \textcolor{red}{$-2$}}   & 74/90   & 20        & 14{\tiny \textcolor{red}{$-6$}}   & 83/97   & 20        & 11{\tiny \textcolor{red}{$-9$}}   \\\midrule
\multirow{3}{*}{\begin{tabular}[c]{@{}l@{}}Close-\\source\end{tabular}}  
& Gemini-3-Flash-Thinking   & 83/94  & 56        & 33{\tiny \textcolor{red}{$-23$}}  & 94/98   & 56        & 35{\tiny \textcolor{red}{$-21$}}  & 98/98   & 56        & 24{\tiny \textcolor{red}{$-32$}}  & 74/90   & 56        & 28{\tiny \textcolor{red}{$-28$}}  & 83/97   & 56        & 24{\tiny \textcolor{red}{$-32$}}  \\
& GPT-5.2-Medium            & 83/94  & 70        & 59{\tiny \textcolor{red}{$-11$}}  & 94/98   & 70        & 55{\tiny \textcolor{red}{$-15$}}  & 98/98   & 70        & 52{\tiny \textcolor{red}{$-18$}}  & 74/90   & 70        & 58{\tiny \textcolor{red}{$-12$}}  & 83/97   & 70        & 51{\tiny \textcolor{red}{$-19$}}  \\
& GPT-5.2-High              & 83/94  & 70        & 70{\tiny \textcolor{gray}{$\pm 0$}} & 94/98   & 70        & 64{\tiny \textcolor{red}{$-6$}}  & 98/98   & 70        & 61{\tiny \textcolor{red}{$-9$}}  & 74/90   & 70        & 67{\tiny \textcolor{red}{$-3$}}   & 83/97   & 70        & 61{\tiny \textcolor{red}{$-9$}}  \\\bottomrule
\end{tabular}}
\label{tab:number_result}
\end{table}

\vspace{-3mm}
\subsection{Result Analysis}
\subsubsection{Solvability and Difficulty}
\begin{wrapfigure}{r}{0.48\columnwidth}
  \centering
  \vspace{-4pt}
  \includegraphics[width=\linewidth]{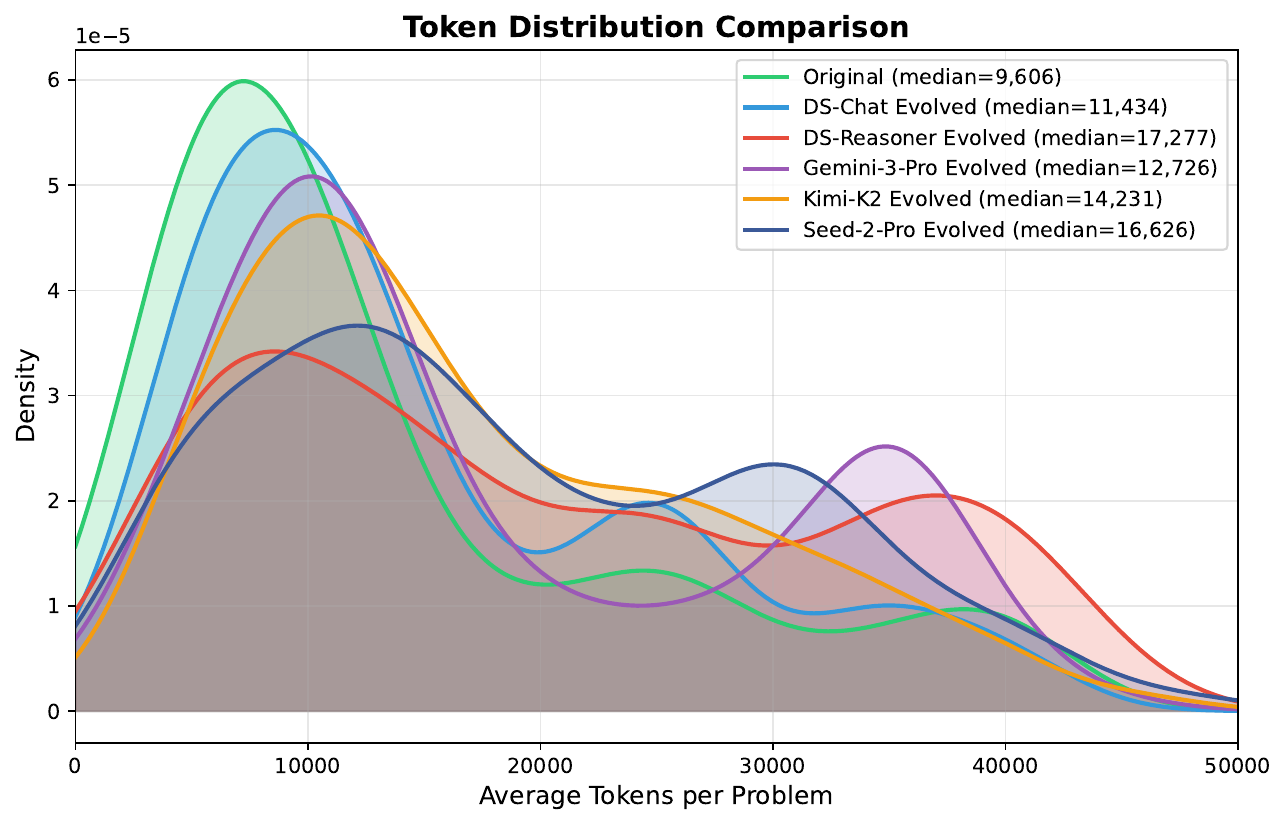}
  \caption{Average token consumption (ATC) for original and evolved problems.}
  \label{fig:distribution}
  \vspace{-6pt}
\end{wrapfigure}
\textbf{Solvability Verification.} 

Table~\ref{tab:number_result} shows high consistency between the internal verifier and the external judge. Across evolution backbones, the external judge certifies between 74 of 90 and 98 of 98 internally accepted problems; \textit{DeepSeek-Reasoner} reaches 94 of 98, and \textit{Gemini-3-Pro-Preview-Thinking} reaches 98 of 98. These results provide a direct check on the reliability of the \textit{Solvability Verification Agent}: although the verifier operates inside the evolving pipeline, most internally accepted generations are also judged valid by the held-out external judge. This suggests that scrutinizing the proposed solution steps for logical flaws is an effective way to filter invalid generations before they enter downstream difficulty evaluation.

\textbf{Difficulty Escalation.} As shown in Table ~\ref{tab:number_result}, most solver--evolver pairs show lower Evolution-SR than Origin-SR, indicating that the evolved problems are systematically harder than the seeds. The effect appears not only on weaker solvers but also on strong closed-source models: \textit{GPT-5.2-High} drops from 70\% to 64\% on \textit{DeepSeek-Reasoner} evolutions and to 61\% on \textit{Gemini-3-Pro-Preview-Thinking} and \textit{Seed-2.0-Pro} evolutions, while \textit{Gemini-3-Flash-Thinking} drops by up to 32 points. These decreases suggest that the evolved problems introduce additional \textit{Burden of Discovery}, forcing solvers to depart from standard solution templates rather than merely handling superficial paraphrases.
Figure~\ref{fig:distribution} shows that evolved problems shift the ATC distribution to the right, with more high-token and timeout cases. This supports the SR results from a different angle: evolved problems require longer reasoning chains and more test-time exploration even when solvers eventually produce an answer. The high-token tail is especially important because it suggests that the added difficulty comes from deeper search and self-correction rather than only from longer wording or heavier arithmetic. We provide 18 detailed case studies in the Appendix \ref{case_study}, illustrating problem evolution across mathematical problems from diverse sources.

\subsubsection{Training Utility of Evolved Problems}
\begin{table}[t]
\centering
\caption{
Training utility of evolved problems.
We report Avg@10 accuracy (\%) at epoch 6 on three AIME-series benchmarks.
\textit{Original} denotes SFT on the original problems only, while
\textit{Original + Evolved} augments the same training set with problems
generated by Code2Math.}
\label{tab:training_utility}
\small
\setlength{\tabcolsep}{5pt}
\begin{tabular}{lccc|ccc}
\toprule
& \multicolumn{3}{c|}{\textbf{Qwen3-8B}}
& \multicolumn{3}{c}{\textbf{Qwen3-14B}} \\
\cmidrule(lr){2-4}\cmidrule(lr){5-7}
Benchmark
& Original & Orig.+Evol. & $\Delta$
& Original & Orig.+Evol. & $\Delta$ \\
\midrule
AIME24 & 42.3 & \textbf{48.0} & +5.7
       & 60.7 & \textbf{66.7} & +6.0 \\
AIME25 & 36.7 & \textbf{38.0} & +1.3
       & 48.0 & \textbf{51.3} & +3.3 \\
AIME26 & 40.7 & \textbf{47.3} & +6.6
       & 58.0 & \textbf{62.7} & +4.7 \\
\midrule
Average & 39.9 & \textbf{44.4} & +4.5
        & 55.6 & \textbf{60.2} & +4.7 \\
\bottomrule
\end{tabular}
\end{table}

Table~\ref{tab:training_utility} evaluates whether the evolved problems provide useful supervision beyond the original training data. We compare SFT on the original problems with SFT on the combination of original and Code2Math-evolved problems, and report Avg@10 accuracy at epoch 6. Augmenting the training set with evolved problems consistently improves performance across all three AIME-series benchmarks and both model scales. For Qwen3-8B, the average accuracy increases from 39.9\% to 44.4\% (+4.5 points), with the largest gain of 6.6 points on AIME26. Qwen3-14B exhibits a similar trend, improving from 55.6\% to 60.2\% (+4.7 points), including gains of 6.0, 3.3, and 4.7 points on AIME24, AIME25, and AIME26, respectively. These consistent improvements suggest that the evolved problems are not merely harder evaluation instances, but also provide additional training signals that benefit downstream mathematical reasoning.

\subsubsection{Efficiency Analysis}



\begin{wraptable}{r}{0.68\columnwidth}
\vspace{-8pt}
\centering
\caption{Average failure counts during problem evolution, decomposed into solvability- and difficulty-verification failures.}
\vspace{-2mm}

\resizebox{\linewidth}{!}{%
\begin{tabular}{lccc}
\toprule
\textbf{Model} & \textbf{Total} & \textbf{Solvability} & \textbf{Difficulty} \\
\midrule
DeepSeek-Chat & 4.11 & 2.65 & 1.46 \\
DeepSeek-Reasoner & 3.10 & 2.08 & 1.02 \\
Gemini-3-Pro-Preview-Thinking & 1.56 & 0.46 & 1.10 \\
Seed-2.0-Pro & 2.57 & 0.78 & 1.79 \\
Kimi-K2-Thinking & 6.55 & 5.71 & 0.84 \\
\bottomrule
\end{tabular}%
}
\label{tab:pe-failure-stats}
\vspace{-8pt}
\end{wraptable}

Table~\ref{tab:pe-failure-stats} shows that producing a qualified evolved problem generally requires multiple attempts, but the efficiency varies substantially across evolution backbone models. 
\textit{Gemini-3-Pro-Preview-Thinking} is the most efficient evolver, with only 1.56 failed rollouts on average, followed by \textit{Seed-2.0-Pro} (2.57) and \textit{DeepSeek-Reasoner} (3.10). 
In contrast, \textit{Kimi-K2-Thinking} requires 6.55 failed rollouts per successful evolution on average, more than four times that of Gemini, indicating that the choice of evolution backbone has a substantial effect on test-time computation.

The decomposition further reveals that different models encounter different bottlenecks. 
For \textit{DeepSeek-Chat}, \textit{DeepSeek-Reasoner}, and especially \textit{Kimi-K2-Thinking}, most failures are rejected by the solvability verifier; for Kimi-K2, 5.71 of the 6.55 average failures arise at this stage. 
This suggests that these models can often propose sufficiently ambitious modifications, but have greater difficulty maintaining a logically consistent problem--solution pair. 
By contrast, \textit{Gemini-3-Pro-Preview-Thinking} and \textit{Seed-2.0-Pro} incur more difficulty-verification failures than solvability failures. 
Their generated candidates are therefore more likely to remain valid, while the remaining challenge is to introduce a sufficiently strong increase in the \textit{Burden of Discovery}. 
This distinction suggests that evolution efficiency is not governed by a single failure mode, but depends on the capability profile of the underlying model.

\begin{figure}[t]   
  \centering
  \includegraphics[width=\linewidth]{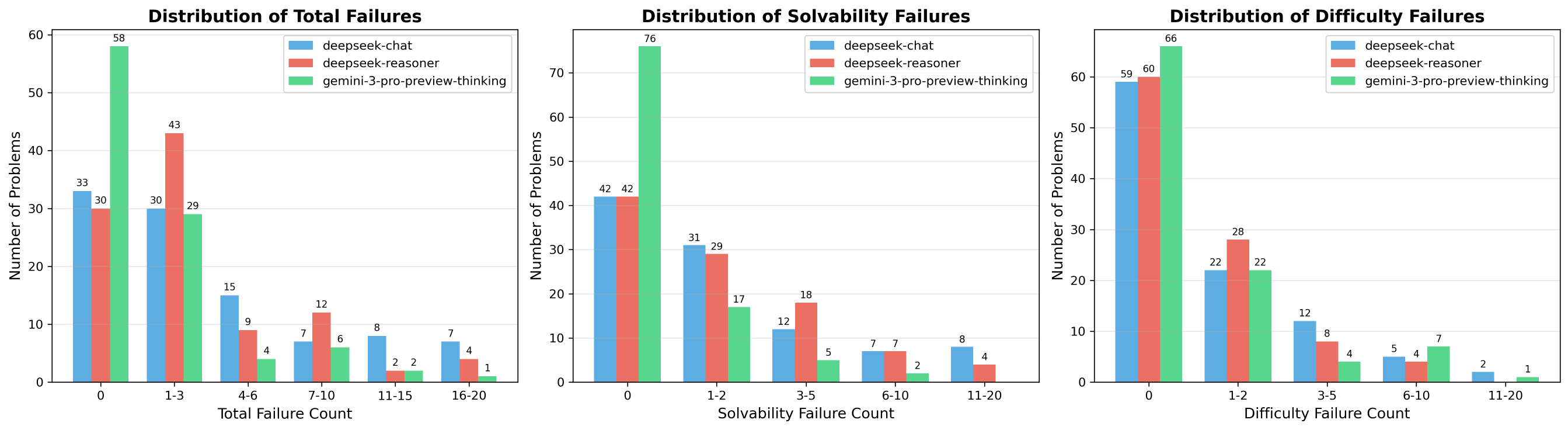}
  \caption{Efficiency Analysis of Agentic Problem Evolution. We visualize the distribution of failure counts encountered during the evolutionary process across three base models: DeepSeek-Chat, DeepSeek-Reasoner, and Gemini-3-Pro-Preview-Thinking. 
  }
  \label{fig:efficiency}
\end{figure}

Figure~\ref{fig:efficiency} further shows that the evolution cost is highly non-uniform across problem instances. 
While many seeds yield an accepted evolution after only a small number of attempts, the distribution exhibits a noticeable long tail in which substantially more exploration is required. 
Such cases indicate that test-time exploration is particularly important for problems whose valid modification space is narrow or for which increasing difficulty without breaking solvability is challenging. 
Overall, these results reveal a central trade-off in Code2Math: dual verification substantially constrains the search toward valid and genuinely harder problems, but doing so requires nontrivial and model-dependent test-time computation~\cite{liu2025costbench}.

\subsection{Human Evaluation}

\textbf{Audit Setup.} Since our automatic evaluation relies on \textit{GPT-5.2-High} as a judge, we conduct a human audit to validate the reliability of these judgments. The audit covers three parts of the evaluation pipeline: evolved-problem validity, solver-answer correctness, and relative difficulty. The main audit is conducted by a postgraduate student in mathematics, using the same evaluation criteria as the automatic prompts. For solver-answer correctness, we binarize scores by mapping $0.5$ to $0$ and use a balanced audit set with equal numbers of GPT-positive and GPT-negative cases.

\textbf{Judge Agreement.} Table~\ref{tab:human-audit-main} shows that GPT judgments align closely with human judgments. Among 127 deduplicated evolved problems judged valid by GPT-5.2-High, humans judge 126 as valid, yielding 99.2\% validity precision. For solver-answer correctness, human and GPT judgments agree on 1134 of 1226 audited responses. For difficulty scoring, exact agreement on the 1--5 scale is 79.3\%, while 95.9\% of examples differ by at most one point, with a small mean bias of $-0.041$. These results suggest that \textit{GPT-5.2-High} provides a reliable automatic judge for the main evaluation signals used in our experiments; full audit statistics are reported in Appendix~\ref{app:human-eval-details}.

\textbf{Olympiad-Level Audit.} We further audit 35 evolved problems whose seeds come from high-level olympiad sources, including IMO shortlist, IMO official, and CMO official problems. Because these examples require stronger mathematical expertise, this subset is additionally reviewed by a mathematics postdoctoral researcher. Under the score $\geq 3$ threshold for successful difficulty improvement, both humans and GPT-5.2-High judge 34 of 35 problems as improved. The average human difficulty score is 4.03, compared with 3.80 from GPT. This indicates that the evolved problems are not merely passing a weak validity filter; even on competition-level seeds, human judgments confirm substantial difficulty gains.

\begin{table}[!htbp]
\centering
\captionsetup{font=footnotesize,justification=raggedright,singlelinecheck=false}
\begin{minipage}[t]{0.50\linewidth}
\vspace{0pt}
\caption{Human audit of GPT-5.2-High judgments.}
\label{tab:human-audit-main}
\scriptsize
\setlength{\tabcolsep}{1pt}
\renewcommand{\arraystretch}{1.05}
\begin{tabular*}{\linewidth}{@{\extracolsep{\fill}}lll@{}}
\toprule
Audit item & Metric & Result \\
\midrule
Validity & Human-valid rate & 126/127 = 99.2\% \\[2.27pt]
Solver corr. & Human--GPT agree. & 1134/1226 = 92.5\% \\[2.27pt]
Difficulty & Exact agree. & 115/145 = 79.3\% \\[2.27pt]
Difficulty & Within-one agree. & 139/145 = 95.9\% \\[2.27pt]
IMO/CMO & Mean score (H/G) & 4.03/3.80 \\[2.27pt]
IMO/CMO & Score $\geq 3$ (H/G) & 34/35; 34/35 \\[2.27pt]
\bottomrule
\end{tabular*}
\end{minipage}\hfill
\begin{minipage}[t]{0.47\linewidth}
\vspace{0pt}
\caption{Failure after evolution (human-valid).}
\label{tab:human-audit-breakdown}
\scriptsize
\setlength{\tabcolsep}{1.5pt}
\renewcommand{\arraystretch}{1.05}
\begin{tabular*}{\linewidth}{@{\extracolsep{\fill}}lrlrr@{}}
\toprule
Solver & Rate & Evolver & Rate & Diff. \\
\midrule
DS-Chat & 74.2\% & Gemini-Pro & 61.8\% & 4.55 \\
Qwen-235B & 64.3\% & Seed & 55.7\% & 3.41 \\
DS-Reasoner & 55.2\% & DS-Reasoner & 50.7\% & 3.62 \\
Gemini-Flash & 55.2\% & DS-Chat & 41.4\% & 3.46 \\
GPT-5.2-M & 40.0\% & Kimi-K2 & 33.9\% & 3.77 \\
GPT-5.2-H & 37.3\% & Overall & 49.1\% & 3.93 \\
\midrule
Overall & 49.1\% &  &  &  \\
\bottomrule
\end{tabular*}
\end{minipage}
\end{table}

\textbf{Failure After Evolution.} Beyond judge agreement, we also test whether evolved problems disrupt solver behavior under human-validated validity. We focus on cases where a solver correctly solves the seed problem but fails on the corresponding evolved problem, restricting the analysis to evolved problems judged valid by both \textit{GPT-5.2-High} and human annotators. This setting directly measures whether evolution breaks the solver's original solution pattern rather than only changing aggregate solve rates.

As shown in Table~\ref{tab:human-audit-breakdown}, this occurs in 166 of 338 applicable solver--problem pairs, a 49.1\% failure-after-evolution rate. Stronger solvers such as \textit{GPT-5.2-High} and \textit{GPT-5.2-Medium} are more robust, while \textit{DeepSeek-Chat} and \textit{Qwen3-235B-A22B-Thinking} are more frequently disrupted. By evolver, \textit{Gemini-3-Pro} produces the highest disruption rate at 61.8\% and also receives the highest average human difficulty score, 4.55.

\section{Limitations}
Our work has several limitations. 
First, our controlled evaluation of problem-evolution quality is conducted on 100 seed problems. 
Although these problems span diverse mathematical domains, sources, and difficulty levels, a substantially larger evaluation set would provide stronger evidence for the generality of our findings. 
Scaling this evaluation is computationally expensive, as each seed requires repeated problem generation, solvability verification, difficulty verification, and downstream solver evaluation.
Second, although our training experiments demonstrate that evolved problems provide useful supervision beyond the original data, our current framework does not yet realize a closed-loop self-improvement process. 
Problem evolution and model training are conducted as separate stages, rather than allowing a model to autonomously generate new problems, learn from them, and then use its improved capabilities to drive subsequent rounds of evolution. 
Third, evaluating the quality of mathematical problems remains inherently difficult and often requires expert judgment. 
While we conduct human audits covering problem validity, solver correctness, relative difficulty, and an additional olympiad-level subset, exhaustive expert evaluation of all evolved problems is impractical. 
Moreover, mathematical problem quality extends beyond solvability and difficulty to dimensions such as novelty, elegance, and pedagogical value, which are not comprehensively assessed in our current evaluation.

\section{Conclusion}
We presented \textsc{Code2Math}, a code-driven multi-agent framework for autonomously evolving mathematical problems through test-time exploration and dual verification. Across multiple evolution and solver models, Code2Math generates mathematically valid problems that are substantially more challenging than their seeds, while augmenting the original training data with evolved problems consistently improves downstream performance across three AIME-series benchmarks. Human evaluation further supports the reliability of these results: expert audits show high agreement with the automatic judgments on problem validity, solver correctness, and relative difficulty, including a dedicated subset of olympiad-level problems. Our analysis also shows that successful evolution requires nontrivial test-time exploration, with different models exhibiting distinct bottlenecks in maintaining solvability or increasing difficulty. These findings suggest that executable problem evolution is useful not only for constructing harder mathematical challenges, but also for producing valuable learning signals for model improvement. A natural next step is to close the loop between evolution and training, allowing improved models to iteratively generate and learn from new problems, moving toward recursively self-improving reasoning systems.

\section*{AI use statement}
We used the web-based GPT as an AI writing assistant to help polish the presentation of the paper, including improving clarity, grammar, and conciseness. This assistant is not used for idea proposing, experimental design or result interpretation. All claims, experimental results, analyses, and final manuscript content were reviewed and verified by the authors. 

\section*{Ethics statement}
This work studies the automatic evolution of mathematical reasoning problems through code-driven agents. 
The main potential risk is that automatically generated problems may contain subtle mathematical errors, misleading solution steps, or apparent difficulty caused by artificial complexity rather than genuine reasoning challenges. 
To mitigate these risks, Code2Math employs dedicated solvability and difficulty verification agents, followed by external model-based evaluation and sampled expert human audits. 
We additionally inspect generated data for malformed or otherwise inappropriate outputs before using them for evaluation or training.

Our experiments use mathematical problems collected from publicly available benchmarks, competition materials, and other research datasets, together with model-generated problem adaptations and solutions. 
We use these artifacts for research on mathematical problem evolution, reasoning evaluation, and model training. 
For both the evaluation and training datasets, users of the released artifacts should respect the licenses, terms of use, and redistribution requirements of the original data sources. 
Generated derivatives should likewise be used in accordance with the access conditions of their source materials. 
The data used in this work consists of mathematical problems, solutions, model outputs, code-execution traces, and evaluation results, and does not contain personal or user-generated records intended to identify individuals.

Our computational experiments involve both provider-supported model APIs and open-weight language models. 
In addition to inference-time problem evolution and evaluation, we conduct supervised fine-tuning experiments to study whether evolved problems provide useful additional training signals. 
Training is performed only on mathematical reasoning data and model-generated solutions; no personal data is used for model training.

Human annotation is used solely for expert evaluation of mathematical problem quality, including validity, solver correctness, and relative difficulty. 
The annotators are authors of this paper with relevant mathematical expertise, including a postgraduate student in mathematics and a postdoctoral researcher. 
They follow predefined annotation criteria aligned with the automatic evaluation protocol, and the corresponding instructions are provided in the appendix. 
No external participants or crowdworkers are recruited, and no personal or sensitive information is collected through the annotation process.

\section*{Reproducibility statement}
The code and experimental resources for Code2Math are publicly available at
\url{https://github.com/TarferSoul/Code2Math}.
The repository includes the implementation of the Code2Math framework, agent
prompts, released problem data, and scripts for running the main problem-evolution
pipeline. We also provide instructions for environment setup, model configuration,
dataset preparation, and execution of the evolution procedure. Additional
experimental details, including the configurations used for downstream evaluation
and training, are documented in the paper.



\bibliographystyle{iclr2027_conference}
\bibliography{custom}
\appendix
\input{appendix}

\end{document}

%% file: math_commands.tex
\usepackage{amsmath,amsfonts,bm}

\def\eqref#1{equation~\ref{#1}}

\def\1{\bm{1}}

\DeclareMathAlphabet{\mathsfit}{\encodingdefault}{\sfdefault}{m}{sl}
\SetMathAlphabet{\mathsfit}{bold}{\encodingdefault}{\sfdefault}{bx}{n}



%% file: appendix.tex
\section{Appendix}
In this appendix, we present additional experimental setup details, the prompts used for the three types of agents, and cases of evolution.

\subsection{Agentic Environment}
\label{app:setup-details}
The multi-agent evolving system is built upon the Smolagents framework~\citep{smolagents}, which enables agents to execute user-defined Python code within a controlled environment. To support complex problem generation and verification, we equip the agents with a comprehensive toolset. This includes standard utility libraries for general-purpose functionality (\textit{json}, \textit{math}, \textit{random}, \textit{statistics}, as well as \textit{itertools} and \textit{collections} for efficient data manipulation). For precision and textual processing, we include \textit{fractions}, \textit{decimal}, \textit{re}, and \textit{functools}. Furthermore, the environment supports advanced scientific and symbolic computing through \textit{numpy}, \textit{scipy}, \textit{pandas}, \textit{openpyxl}, \textit{sympy}, \textit{mpmath}, and \textit{z3}, alongside \textit{networkx} for graph operations.

\subsection{Why Code Helps}
\label{app:code_helps}

To understand how executable exploration contributes to problem evolution, we analyze 646 Python calls collected from 100 Gemini evolution trajectories and categorize each call by its primary function. As shown in Table~\ref{tab:code_function_analysis}, code serves substantially broader purposes than post-hoc correctness checking. Construction search, counterexample pruning, and parameter search together account for 43.3\% of all executions, exceeding verification and edge-case checking, which constitute 34.2\%. These design-oriented operations allow the agent to enumerate candidate structures, eliminate invalid conjectures, and search for constants or thresholds that can directly determine the formulation of an evolved problem.
The remaining executions further support mathematical exploration through pattern discovery, numerical experimentation, specialized solvers, graph algorithms, dynamic programming, and symbolic derivation. In contrast, verification and edge-case analysis provide executable feedback for testing intermediate hypotheses and rejecting invalid directions before they are incorporated into the final problem. Overall, the distribution reveals two complementary roles of code: it acts both as a search mechanism for constructing new mathematical structures and as a verification mechanism for filtering candidate constructions. This suggests that executable environments contribute to problem evolution not merely by checking proposed solutions, but by actively shaping the search process through which new problems are formed.
\begin{center}
\captionof{table}{Functional breakdown of Python executions in 100 Gemini evolution trajectories.}
\label{tab:code_function_analysis}
\footnotesize
\setlength{\tabcolsep}{5pt}
\begin{tabular*}{0.78\columnwidth}{@{\extracolsep{\fill}}lc}
\toprule
Function & Calls (\%) \\
\midrule
Verification / edge cases & 221 (34.2) \\
Construction search & 154 (23.8) \\
Counterexample pruning & 67 (10.4) \\
Parameter search & 59 (9.1) \\
Pattern discovery & 42 (6.5) \\
Solver / graph / DP & 40 (6.2) \\
Numerical experiments & 32 (5.0) \\
Symbolic derivation & 31 (4.8) \\
\midrule
\textbf{Total} & \textbf{646 (100.0)} \\
\bottomrule
\end{tabular*}
\vspace{-3mm}
\end{center}

\subsection{Additional Human Evaluation Details}
\label{app:human-eval-details}

\begin{table}[h]
\centering
\caption{Full human audit of GPT-5.2-High judgments. The upper block measures agreement between human annotators and GPT-5.2-High; the lower block focuses on difficulty improvement for IMO/CMO-source problems.}
\label{tab:human-audit-full}
\scriptsize
\setlength{\tabcolsep}{5pt}
\begin{tabular}{lllc}
\toprule
Audit type & Scope & Metric & Result \\
\midrule
\multicolumn{4}{l}{\textit{Agreement with human judgments}} \\
Validity & GPT-valid evolved problems & Human-valid rate & 126 / 127 = 99.2\% \\
Solver correctness & Balanced solver responses & Human--GPT agreement & 1134 / 1226 = 92.5\% \\
Difficulty score & Clean evolved problems & Exact agreement & 115 / 145 = 79.3\% \\
Difficulty score & Clean evolved problems & Within-one agreement & 139 / 145 = 95.9\% \\
Difficulty score & Clean evolved problems & MAE / RMSE & 0.262 / 0.632 \\
Difficulty score & Clean evolved problems & Human $-$ GPT bias & $-0.041$ \\
Difficulty score & Clean evolved problems & Spearman / QWK & 0.714 / 0.632 \\
\midrule
\multicolumn{4}{l}{\textit{Difficulty improvement on IMO/CMO-source problems}} \\
Difficulty score & IMO/CMO subset & Number of audited problems & 35 \\
Difficulty score & IMO/CMO subset & Mean score, Human / GPT & 4.03 / 3.80 \\
Difficulty pass & IMO/CMO subset & Score $\geq 3$, Human / GPT & 34 / 35 = 97.1\% \quad / \quad 34 / 35 = 97.1\% \\
Strong improvement & IMO/CMO subset & Score $\geq 4$, Human / GPT & 27 / 35 = 77.1\% \quad / \quad 26 / 35 = 74.3\% \\
Top score & IMO/CMO subset & Score $=5$, Human / GPT & 10 / 35 = 28.6\% \quad / \quad 3 / 35 = 8.6\% \\
Difficulty pass & IMO/CMO subset & Pass/fail agreement & 33 / 35 = 94.3\% \\
\bottomrule
\end{tabular}
\end{table}

\subsection{Prompt Templates}
Figure \ref{fig:prompt_1} to \ref{fig:prompt_6} present the detailed prompts used in Code2Math, including those for the three agents in our framework as well as the prompts used to judge answer correctness, problem solvability, and related evaluation criteria.
\begin{figure}[!ht]
\begin{tcolorbox}[
    colback=gray!3!white,
    colframe=black!30!white, 
    title=Evolution Agent Prompt Template,
    fonttitle=\bfseries,
    boxrule=0.5pt,
    arc=4pt,
    boxsep=5pt,
    left=6pt,
    right=6pt,
    top=6pt,
    bottom=6pt,
    coltitle=blue!20!black,
    fontupper=\tiny,
    fontlower=\tiny
]
Part 1: Your Mission and Core Principles

You are an expert agent specializing in Mathematical Problem Adaptation. Your task is to analyze a given problem and solution, identify its primary bottleneck (what makes it difficult), and the core mathematical insight required to solve it. Using this analysis, you will then formulate a novel, higher-order problem of substantially greater difficulty, and provide a comprehensive, step-by-step solution for the new problem.

Part 2: Your Working Process

Your entire process is a continuous, step-by-step cycle. In each round, analyze the original problem and think carefully about potential directions for adapting the problem, then write code (for example, using sympy for symbolic math) to explore and validate those directions.
- Code is your whiteboard for performing symbolic algebra, testing numeric cases, or verifying properties needed to construct the new problem.
- After your code is executed, you will receive deterministic output (for example, a simplified expression or a numerical result).

Part 3: Critical Mandates (Non-Negotiable Process Rules)

Workflow Mandate: Exploration First, Final Answer Last (VERY IMPORTANT)

Your process has two phases:

1) Exploration Phase
Move beyond simple derivation. First, adopt the mindset of an elite mathematical competitor and analyze the provided solution to identify the original problem’s true bottleneck. Locate the precise conceptual hurdle or non-obvious starting point that causes difficulty. After isolating this core challenge, engineer a more formidable obstacle. Either escalate the existing bottleneck or design a new, related one, ensuring that the path to the solution is significantly more obscured. This design process must be iterative, using deep mathematical knowledge and computational validation with Python libraries to confirm the integrity and heightened difficulty of your construct.
Examples of valued insights:
- Combinatorics: A delicate construction or a clever bijective argument based on keen observation, not just a standard formula.
- Sequences or Number Theory: A subtle underlying pattern, hidden periodicity, or a law governing the distribution of terms.
- Analysis: A key qualitative property of a function (such as symmetry, bounds, or geometric meaning of its derivative) that standard procedures would overlook.

2) Finalization Phase
This is your very last action. After completing derivations and being confident in your new problem, call the final\_answer tool with a correctly formatted Python dictionary.

Part 4: Guiding Principles of Mathematical Construction (Content Rules)

1) The Golden Rule of Problem Design: The Burden of Discovery and Insight (CRITICAL PRINCIPLE)
Maximize conceptual difficulty to force a hard-won “Eureka” moment. The adapted problem must be constructed so that even a competition-level solver struggles to find the entry point. The solution should only be reachable after extensive observation, experimentation, and trial to unearth the deep insight. Avoid superficial difficulty such as heavier calculations or more procedural steps; the challenge must originate from the intellectual leap required to begin.

2) Principle of Logical Integrity and Solvability (CRITICAL PRINCIPLE)
Ensure the constructed problem is well-defined, solvable, and unambiguous. Clearly state conditions and constraints. Aim for challenging but fair puzzles.

3) New Problem Categories and Answer Formatting (CRITICAL PRINCIPLE)
Create challenging math problems in one of the following categories. Match the output format to the problem type.

Category 1: Definitive Answer Problems (Calculation or Derivation)
- Accepted answer types: number, simplified algebraic expression, function, interval or set of numbers.
- Output format:
  - “new\_problem”: the problem statement.
  - “new\_solution\_steps”: the full derivation and step-by-step logic.
  - “new\_answer”: the final computed result.
- If the answer format could be ambiguous, specify it in “new\_problem” (for example, “give your answer as a fraction in simplest terms” or “express the function in polynomial form”).

Category 2: Proof-Based Problems
- Nature: ask the solver to prove a mathematical statement.
- Output format:
  - “new\_problem”: the statement to prove.
  - “new\_solution\_steps”: a complete, rigorous, and logically sound proof.
  - “new\_answer”: None.

Category 3: Algorithm Design Problems
- Nature: ask the solver to design an algorithm meeting specific constraints.
- Output format:
  - “new\_problem”: the algorithm design task and constraints.
  - “new\_solution\_steps”: a clear description of the algorithm, proof of correctness, and time/space complexity analysis.
  - “new\_answer”: a concise description of the algorithm’s output if applicable.

When adapting, preserve the fundamental nature of the original problem. If the original required a numerical result, the adapted problem should also have an answer that is simple to verify. For proof or algorithm design problems, maintain the original format.

Part 5: Final Output Specification

Final Output Format (VERY IMPORTANT)
After exploration, your final action is to call the final\_answer tool. Its argument must be a single, valid Python dictionary with exactly three keys: “new\_problem”, “new\_solution\_steps”, and “new\_answer”.

Key Descriptions
- “new\_problem”: a clear, complete, self-contained description of the evolved problem.
- “new\_solution\_steps”: a numbered, human-readable summary of the key logical steps for the solution (derivation, proof, or algorithm).
- “new\_answer”: for definitive-result problems, a string representing the value; for proof or algorithm problems, use the Python literal None.

Part 6: Available Resources

Examples: {demonstrations}. These show adaptations of math problems and illustrate the workflow and baseline for increasing difficulty. In practice, you must rigorously explore the subject and elevate complexity to the maximum possible extent. Do not oversimplify.

Part 7: Final Checklist

1) Strictly follow all formatting requirements.
2) You may draw inspiration from high-level competitions (e.g., IMO), but do not copy or superficially adapt known problems.

Now begin!

\end{tcolorbox}
\caption{The prompt template of our \textit{Evolution Agent}.}
\label{fig:prompt_1}
\end{figure}

\begin{figure}[h!]
\begin{tcolorbox}[
    colback=gray!3!white,
    colframe=black!30!white, 
    title=Solvability Verification Agent Prompt Template,
    fonttitle=\bfseries,
    boxrule=0.5pt,
    arc=4pt,
    boxsep=5pt,
    left=6pt,
    right=6pt,
    top=6pt,
    bottom=6pt,
    coltitle=blue!20!black,
    fontupper=\tiny,
    fontlower=\tiny
]

Part 1: Your Role \& Mission
You are the Lead Mathematical Solvability Auditor. Your sole purpose is to ruthlessly stress-test a given mathematical problem to ensure it is logically sound, non-contradictory, and correctly solvable.
You act as a firewall against "bad math". You do not care if the problem is interesting or hard; you only care if it is broken.
You will receive:
1. \`problem\_text\`: The statement of the new problem.
2. \`proposed\_solution\`: The step-by-step derivation provided by the creator.
3. \`answer\`: The expected final result.
   - Note: This is NOT limited to numerical values. It can be an algebraic expression, a tuple, a set of values, or a function.
   - For Proof Problems: If the problem asks to "Prove that..." or "Show that...", this field may be \`None\`, \`null\`, or a placeholder string (e.g., "N/A"). In these cases, the "answer" is considered the successful completion of the logical argument in the solution steps.
Your work is separated into two distinct phases. You must pass Phase 1 (The Static Check) before moving to Phase 2 (The Logic Audit).

Part 2: Phase 1 - Static Problem Analysis (The "Sanity Check")
Before looking at the solution steps, you must validate the \`problem\_text\` itself. You are looking for internal inconsistencies or definitions that are mathematically illegal.
Verification Checklist for Phase 1:
1. Domain \& Value Rationality:
   - Are all constants and variables within valid domains? (e.g., denominators \(\neq\) 0, even roots of negatives, \(\arcsin(x)\) where \(|x|>1\), logarithms of non-positive numbers).
   - Are the physical/geometric values inherently possible? (e.g., Probability \(P \in [0,1]\), Triangle sides satisfy \(a+b>c\), Friction coefficient \(\mu > 0\), Mass \(m \geq 0\)).
2. Constraint Consistency (Over-definition Check):
   - Does the problem provide too many conditions that contradict each other?
   - Action: You MUST attempt to model the problem's geometric or algebraic constraints in Python (e.g., using \`sympy\` or geometric coordinate geometry). If the constraints lead to an empty set (e.g., "Find a real number \(x\) such that \(x > 5\) and \(x < 3\)"), the problem is INVALID.
If the problem fails Phase 1 (contains illegal definitions or contradictions), you stop immediately and report the error.

Part 3: Phase 2 - Step-by-Step Logic Audit (The "Deep Dive")
If the problem text is valid, you proceed to audit the \`proposed\_solution\`. You must verify each step \(S\_1, S\_2, ... S\_n\) extensively.
For Every Step, Perform These 3 Checks:
1. Conflict Check: Does the intermediate conclusion obtained in this step contradict the original \`problem\_text\`? (e.g., The problem says \(x\) is an integer, but this step derives \(x = 0.5\)).
2. Derivation Verification (The Code Audit):
   - Do not trust the text. Use Python to independently calculate the transformation from Step \(N\) to Step \(N+1\).
   - Validating an integral? Use \`sympy.integrate\`. Solving an equation? Use \`sympy.solve\`.
3. Logical Fallacy Detection: check if the step commits any of the following specific errors.
Formal Failure Modes (The "Red Flags"):
- [Transformation Error]: Recasting a target statement into a non-equivalent or strictly weaker one (e.g., proving \(A \Rightarrow B\) when \(A \iff B\) was required).
- [Over Generalization]: Drawing a universal conclusion from special cases (e.g., verifying for \(n=1,2,3\) and assuming for all \(n\)).
- [Invalid Construction]: Introducing an object that cannot exist (e.g., "Let \(f(x)\) be a polynomial with infinite roots").
- [Wrong Division]: Case analysis that misses possibilities (e.g., checking \(x>0\) and \(x<0\) but forgetting \(x=0\)).
- [Circular Reasoning]: Using the conclusion as a hidden premise.
- [Logic Violation]: Algebraic illegal moves (e.g., dividing by a variable that could be zero).
- [Hidden Assumption]: Using a theorem without verifying its preconditions (e.g., applying L'Hopital's rule without checking \(0/0\) form).
- [Boundary Neglect]: Ignoring edge cases in optimization or integration limits.
- [Vague Argument]: Using "obviously" or hand-waving instead of rigorous derivation.
Final Holistic Review (The "Verdict"):
- Determine whether there are failure modes in the overall logical chain.
- Determine whether the final logical conclusion actually answers the specific question asked in \`problem\_text\`.

Part 4: Your Interactive Workflow: A Multi-Turn Process
Your entire process is a continuous, step-by-step cycle. In each round, you should think carefully, then write code (e.g., using \`sympy\` for symbolic math) to validate your thoughts.
- Remember code is your whiteboard for performing symbolic algebra, testing numeric cases, or verifying properties.
- After your code is executed, you will receive the deterministic output of your code (e.g., a simplified expression, a numerical result).

Part 5: Final Output Specification
Your final output must be a single call to the \`final\_answer\` tool. The only argument must be a Python dictionary with exactly two keys: \`"status"\` and \`"reason"\`.
Example of a final answer (Passing Scenario):
{
    "status": "PASS",
    "reason": "The problem text is rigorous and self-consistent. The proposed solution's derivation was verified step-by-step using Python/SymPy. The logic chain is complete and accurate."
}
Example of a final answer (Failing Scenario):
{
    "status": "FAIL",
    "reason": "The solution fails at Step 4 due to a [Wrong Division] error. The logical argument assumes x > 0, but the problem domain allows for x = 0, which leads to a singularity. Furthermore, the Global check reveals a sufficiency failure; the final answer includes a value that satisfies the derived equation but violates the initial geometric constraints."
}

Part 6: Notes
1. Your final output must be a single call to the \`final\_answer\` tool.
2. Be meticulous and objective. Your role is a strict verifier. Any discrepancy must be reported and result in a failure.
Now begin your audit.

\end{tcolorbox}
\caption{The prompt template of our \textit{Solvability Verification Agent}.}
\label{fig:prompt_2}
\end{figure}

\begin{figure}[h!]
\begin{tcolorbox}[
    colback=gray!3!white,
    colframe=black!30!white, 
    title=Difficulty Verification Agent Prompt Template,
    fonttitle=\bfseries,
    boxrule=0.5pt,
    arc=4pt,
    boxsep=5pt,
    left=6pt,
    right=6pt,
    top=6pt,
    bottom=6pt,
    coltitle=blue!20!black,
    fontupper=\tiny,
    fontlower=\tiny
]
Part 1: Your Mission \& Role
You are a Specialist in Mathematical Problem Difficulty Assessment. Your mission is to determine if a new\_problem represents a significant and elegant leap in difficulty compared to an original\_question, warranting a PASS status.
You must embody the mindset of an experienced mathematician and educator. Your evaluation should confirm that the adapted problem's difficulty is elevated in terms of conceptual depth and mathematical insight, rather than simply being made more tedious through increased computational complexity or longer but straightforward procedures. Your judgment is about whether the path to its solution requires a fundamentally deeper level of thinking. A minor increase in difficulty is insufficient and must result in a FAIL.
You will be provided with the original problem and its solution, followed by the new problem and its solution. You are to assume the provided solutions are mathematically correct. Your exclusive focus is on comparing the required problem-solving methodologies.

Part 2: The Core Difficulty Assessment Criteria
To receive a PASS, the new task must satisfy the following high-level conditions:
The Nature of the Increased Difficulty: Insight over Execution
The difficulty increase must stem from a higher-order cognitive demand, not merely from increased computational labor. You must strictly filter out "trivial" difficulty increases, which automatically lead to a FAIL.
- INVALID (FAIL) Difficulty Increases:
  - Computational Inflation: Using larger numbers, more complex functions that don't change the underlying logic, or requiring more steps of a known algorithm.
  - Variable Substitution: Simply changing variables or the presentation format without altering the core solution strategy.
  - Minor Twists: Adding a simple, straightforward condition that is easily handled by a small modification to the original method.
- VALID (Required for PASS) Difficulty Increases:
  - The adapted problem actively resists solution by common templates or straightforward, algorithmic approaches. Its solution path should not be immediately apparent, forcing the solver beyond mere pattern-matching or procedural execution.
    For clarification, consider a problem that a student in a standard curriculum would find very challenging, but which a student trained for math Olympiads would instantly recognize as a standard problem type solvable by a learned trick. This is a form of pseudo-difficulty based on specialized training, not the universal conceptual depth we seek, and should be avoided.
  - The solution of the new problem requires a Eureka moment or a non-obvious insight. This insight should either be entirely new to the problem or represent a significantly deeper, elegant and more sophisticated application of the insight required for the original problem.
    To be specific, we highly value insights that stem from a genuine mathematical discovery. Favored examples include:
      - In Combinatorics: Devising a delicate construction or a clever bijective argument based on keen observation, rather than just applying a standard formula.
      - In Sequences or Number Theory: Uncovering a subtle underlying pattern, a hidden periodicity, or a law governing the distribution of terms.
      - In Analysis: Grasping a key qualitative property of a function (e.g., its symmetry, bounds, or the geometric implication of its derivative) that standard procedures would overlook.
  - The adapted problem, in its statement or conclusion, represents a clear mathematical escalation from the original. This means the new problem is not just different, but fundamentally deeper. We specifically value adaptations that achieve one of the following:
      - Generalize the Original Result: The new problem asks to prove a broader theorem, for which the original problem's result is merely a specific instance or a stepping stone.
      - Optimize a Condition: The new problem seeks a provably tighter bound or an exact, optimal constant, where the original might have only asked for a simpler inequality or estimate.
      - Refine the Constraints: The new problem introduces a more subtle or elegant set of constraints that fundamentally alters the problem's landscape, demanding a more sophisticated understanding to even begin the analysis.
In essence, these three criteria work in concert to substantially elevate the burden of discovery. The adapted problem should be constructed such that its entry point or the key idea is deliberately obscured. A successful adaptation is one where the solver's primary struggle is not with the complexity of computation or the length of the deduction, but with the profound, creative challenge of finding that first crucial insight.

Part 3: Your Analytical Workflow
Your task is a purely intellectual process of comparison and judgment, culminating in a single, final tool call.
1. Analyze the Original: First, deeply understand the original\_question and its original\_solution\_steps. Classify its difficulty and the core insight required to solve it.
2. Analyze the New: Next, analyze the new\_problem and its new\_solution\_steps. Deconstruct the argument to pinpoint the crucial logical steps and insights required.
3. Compare and Contrast: Directly compare the methodologies. Is the new method just a more laborious version of the old one, or is it fundamentally different? Does it fit the Difficulty Assessment Criteria?
4. Formulate Judgment: Based on the comparison, make a final decision. Does the new problem represent a significant and elegant leap in difficulty, or is the increase minor/computational? This decision directly determines the PASS/FAIL status.
5. Final Output: Once your judgment is formed, proceed directly to the final parts to call the final\_answer tool.
To calibrate your assessment, you must adopt a specific persona: imagine you are evaluating the problem from the perspective of a skilled and experienced competitor in mathematical olympiads.
This is not a novice. This individual has a robust toolkit of standard theorems, inequalities, and problem-solving heuristics. They can quickly identify common patterns and apply standard techniques.
Therefore, when you analyze the adapted problem, ask yourself these critical questions:
- Would this problem force such a competitor to pause and think? Or would the solution path be immediately obvious to them?
- Are their go-to, standard techniques (e.g., a straightforward application of AM-GM, pigeonhole principle, or modular arithmetic) insufficient or intentionally misleading here?
- Does the problem present a genuine, non-trivial challenge that would give even this skilled solver a sense of accomplishment upon finding the solution?
A problem that is easily dispatched by this benchmark competitor, even if difficult for a layperson, has failed to create a sufficient burden of discovery and should be judged accordingly.

Part 4: Your Interactive Workflow: A Multi-Turn Process
Your entire process is a continuous, step-by-step cycle. In each round, you should think carefully, then write code (e.g., using sympy for symbolic math) to validate your thoughts.
- Remember code is your whiteboard for performing symbolic algebra, testing numeric cases, or verifying properties.
- After your code is executed, you will receive the deterministic output of your code (e.g., a simplified expression, a numerical result).

Part 5: Final Output Specification
Your final output must be a single call to the final\_answer tool. The only argument must be a Python dictionary with exactly three keys: "status", "score", and "reason".
- "score": Must be an integer from 1 to 5. You will determine this score based on the scoring rubric below.
- "status": Must be one of two exact string values: "PASS" or "FAIL". This value is strictly determined by the score:
  - If score is 3, 4, or 5, status must be "PASS".
  - If score is 1 or 2, status must be "FAIL".
- "reason": A detailed string of text explaining why you assigned that specific score, referencing the rubric.
You will use the following criteria to score the quality of the adaptation. Your reason text must justify your choice of score.
- Score 1 (FAIL - Unacceptable):
  - Fails to change the core solution path of the original problem.
  - May even lower the difficulty by removing key constraints or adding unhelpful but trivial information.
  - The burden of discovery is unchanged or reduced.
- Score 2 (FAIL - Poor Adaptation):
  - The difficulty is increased, but only superficially.
  - This increase comes from increased computational complexity (e.g., solving a messier polynomial) or more procedural steps (e.g., applying the same simple idea three times instead of once).
  - It does not require any new, profound insight. A skilled solver would find it "tedious," not "difficult." The burden of discovery is not meaningfully increased.
- Score 3 (PASS - Acceptable / Borderline):
  - A competent adaptation that successfully increases the burden of discovery.
  - It manages to break the standard templates required for the original problem, forcing the solver to pause and think of a new angle.
  - The adaptation might lack elegance, or the new Eureka insight might be relatively straightforward for a top competitor, but it meets the minimum requirement of creating a non-trivial obstacle.
- Score 4 (PASS - Excellent):
  - A high-quality, impressive adaptation that fully aligns with our goals.
  - Anti-Templating: Renders standard solution methods ineffective or turns them into traps.
  - Requires Eureka: The solution depends on one or more non-trivial, insightful "Aha!" moments.
  - Mathematical Advancement: The adapted problem is more mathematically interesting—perhaps by connecting different fields, refining constraints, or representing a more general, profound statement.
  - It provides a genuine challenge and a sense of accomplishment for the skilled competitor.
- Score 5 (PASS - Exemplary / Perfect):
  - Meets all the criteria for a score of 4, but additionally possesses a striking quality of mathematical beauty.
  - This could be found in the problem's surprising simplicity, its deep symmetry, an unexpected connection between disparate fields, or its power as a toy model that illustrates a grander concept.
  - The Eureka moment is not just a key to the solution but also a source of aesthetic satisfaction and a moment of genuine mathematical enlightenment for the solver. This is an adaptation worthy of a textbook.

\end{tcolorbox}
\caption{The prompt template of our \textit{Difficulty Verification Agent}.}
\label{fig:prompt_3}
\end{figure}

\begin{figure}[h!]
\begin{tcolorbox}[
    colback=gray!3!white,
    colframe=black!30!white, 
    title=Solvability Evaluator Prompt Template,
    fonttitle=\bfseries,
    boxrule=0.5pt,
    arc=4pt,
    boxsep=5pt,
    left=6pt,
    right=6pt,
    top=6pt,
    bottom=6pt,
    coltitle=blue!20!black
]
Your Task:

1. Check if the problem statement has any logical errors, contradictions, or is ill-defined.

2. Check if the problem is mathematically solvable. Examine the provided reference solution. If the solution is correct as is, or if it can be modified and supplemented to solve the problem, then the problem is considered solvable. Be lenient with the solution; rough drafts with errors are acceptable if they can be corrected to successfully solve the problem.

Response Format (JSON):

\{

  "has\_logic\_error": true/false,
  
  "logic\_error\_description": "description of logic error if any, or null",
  
  "is\_solvable": true/false,
  
  "solution\_correct": true/false,
  
  "solution\_issues": ["list of issues in solution, empty if correct"],

  "overall\_valid": true/false,
  
  "reason": "brief explanation of your verdict"
  
\}

Important:

- A problem is valid if it has no logic errors and is solvable.

- Small bugs in the reference solution (calculation errors, typos) should not invalidate the problem.

- Only mark solution\_correct as false if there are fundamental errors in the approach.

- Respond with only the JSON, no additional text.

\end{tcolorbox}
\caption{The prompt template of the solvability evaluator.}
\label{fig:prompt_4}
\end{figure}

\begin{figure}[h!]
\begin{tcolorbox}[
    colback=gray!3!white,
    colframe=black!30!white, 
    title=Solver Prompt Template,
    fonttitle=\bfseries,
    boxrule=0.5pt,
    arc=4pt,
    boxsep=5pt,
    left=6pt,
    right=6pt,
    top=6pt,
    bottom=6pt,
    coltitle=blue!20!black
]
Role:

You are a distinguished mathematics expert with a strong academic background and a Gold Medal Math Olympiad coach. You excel at solving complex problems in algebra, geometry, calculus, and statistics, placing the highest importance on logical rigor and step-by-step clarity.

Task:

Solve the mathematical problem provided by the user.

Guidelines:

1. Deep Reasoning: Before providing the final answer, you must break down the problem and derive each step in detail.

2. LaTeX Formatting: All mathematical formulas and variables must be written in LaTeX format (e.g., $x^2 + y^2 = z^2$).

3. Output Format: Output the result strictly as a valid JSON object based on the schema below. Do not output any conversational text, explanations, or Markdown outside of the JSON object.

Output JSON Schema:

\{

  "question\_summary": "A brief summary of the problem",
  
  "solution\_steps": [
    {
      "step\_number": 1,
      "description": "Detailed text explanation of this step",
      "calculation": "Key calculations or formulas involved in this step (in LaTeX)"
    },
    {
      "step\_number": 2,
      "description": "...",
      "calculation": "..."
    }
  ],
  
  "final\_answer": "The final concise answer (numerical value or expression in LaTeX)"
  
\}

\end{tcolorbox}
\caption{The prompt template of the problem solver.}
\label{fig:prompt_5}
\end{figure}

\begin{figure}[h!]
\begin{tcolorbox}[
    colback=gray!3!white,
    colframe=black!30!white, 
    title=Solution Evaluator Prompt Template,
    fonttitle=\bfseries,
    boxrule=0.5pt,
    arc=4pt,
    boxsep=5pt,
    left=6pt,
    right=6pt,
    top=6pt,
    bottom=6pt,
    coltitle=blue!20!black
]
Role:

You are a strict and precise Mathematics Examiner. Your task is to evaluate a student's solution to a given math problem based strictly on the validity of their logic and result.

Input Data:

- **Problem**: {{problem}}
- **Student's Solution Steps**: {{solution\_steps}}
- **Student's Final Answer**: {{final\_answer}}

Evaluation Rules

1. **Logic Check**: Review the student's solution steps. Determine if the mathematical derivation is logically sound and mathematically valid.

2. **Consistency Check**: Verify that the final answer naturally follows from the steps provided.

3. **Scoring**: Assign a score based strictly on the rubric below.

Scoring Rubric

- **1.0 (Perfect)**:

  - The reasoning is logically sound and complete.
  
  - The calculation steps are error-free.
  
  - The final answer is correct.
  
- **0.5 (Minor Flaws)**:

  - The final answer is **correct**.
  
  - The core reasoning logic is correct.
  
  - BUT there are minor presentation errors, non-standard notation, or slightly skipped trivial steps that do not affect the validity of the result.
  
- **0.0 (Incorrect)**:

  - The final answer is **incorrect**.
  
  - OR the final answer is correct but derived from fundamentally wrong logic (lucky guess).
  
  - OR the solution is completely missing or irrelevant.

Output Format

Output **only** a valid JSON object. Do not output any markdown code blocks or additional text.

\{

  "score": number, either 0, 0.5, or 1,
  
  "reason": "string, a concise explanation of why this score was given, citing specific errors or confirming correctness"
  
\}

\end{tcolorbox}
\caption{The prompt template of the solution evaluator.}
\label{fig:prompt_6}
\end{figure}

\subsection{Case Study}
\label{case_study}
We presents 18 detailed evolution cases in this section, illustrating how Code2Math adapts mathematical problems from diverse sources, including AIME, the IMO, and the Chinese Gaokao.

\begin{figure}[t]
\begin{showcase}[colframe=red!50!black]{Case 1}
\underline{\textbf{Evolver:}} Gemini-3-Pro-Preview-Thinking\\

\underline{\textbf{Original Problem:}} \\
Given two distinct points \(P(x_1, y_1)\) and \(Q(x_2, y_2)\) on the quadratic function \(y = x^2 + bx - \frac{1}{4}\), if both points \(P\) and \(Q\) are on the left side of the axis of symmetry, and \(x_1\) and \(x_2\) are integers with \(x_1 < x_2\), prove that \(x_1 - x_2 + y_1 - y_2 > 0\).\\

\underline{\textbf{Evolved Problem:}} \\
Let \( f_n(x) = x^3 + n x^2 \) for a positive integer \( n \). Let \( S_n \) be the set of all triangles \( 	riangle ABC \) such that the vertices \( A, B, C \) are distinct points on the graph of \( y = f_n(x) \) and their \( x \)-coordinates are integers strictly less than the inflection point of the graph. Let \( \Delta_n \) be the minimum possible area of such a triangle in \( S_n \). Calculate the value of \( \sum_{n=1}^{2025} \Delta_n \). \\

\underline{\textbf{Comment:}} \\
The adaptation creates an exemplary problem that transforms a specific inequality about integers on a quadratic curve into a rich geometric-algebraic optimization problem on a cubic curve. 

Justification:

1.  **Mathematical Depth (Score 5 Criteria):** The problem creates a deep connection between the geometry of the cubic curve (inflection point, collinearity condition) and number theory. The 'Eureka' moment is realizing that the area of a triangle on the curve $y=x^3+nx^2$ factorizes into a product of coordinate differences and a linear term $(x_1+x_2+x_3+n)$. This mirrors the collinearity condition ($x_1+x_2+x_3 = -n$) and generalizes the original problem's reliance on $x_1+x_2$ vs $-b$.

2.  **Anti-Templating:** The solution path is unique. It requires deriving the specific area determinant, recognizing the product-minimization implies consecutive integers, and then solving the modular arithmetic constraints imposed by the inflection point. It is far from a standard template.

3.  **Elegant Complexity:** Unlike a mere computational increase, the move to a cubic curve introduces the inflection point as the natural boundary (analogous to the axis of symmetry), making the constraint intrinsic to the function's geometry. The resulting periodic sequence (5, 4, 6) is aesthetically satisfying."

\end{showcase}
\end{figure}
\newpage
\begin{figure}[t]
\begin{showcase}[colframe=red!50!black]{Case 2}
\underline{\textbf{Evolver:}} Gemini-3-Pro-Preview-Thinking\\

\underline{\textbf{Original Problem (From IMO 2025):}} \\
Let \( n \geq 3 \) be a given integer. A line in the plane is called **sunny** if it is not parallel to the \(x\)-axis, the \(y\)-axis, or the line \(x + y = 0\).

Consider the set of points \( S_n = \{ (a, b) \in \mathbb{Z}^+ \mid a + b \leq n + 1 \} \). We want to cover all points in \( S_n \) using exactly \( n \) distinct lines.

Determine all possible values of \( k \) (where \( k \) is a non-negative integer) such that there exists a configuration of \( n \) lines covering \( S_n \) with exactly \( k \) sunny lines.\\

\underline{\textbf{Evolved Problem:}} \\
Let \( n \ge 1 \) be an integer. Consider the set of points \( S_n = \{ (x, y) \in \mathbb{Z}^2 \mid x \ge 1, y \ge 1, x + y \le n + 1 \} \). We wish to cover all points in \( S_n \) using exactly \( n \) lines in the Euclidean plane, such that no two lines are parallel (i.e., all \( n \) lines have pairwise distinct slopes).

Find the largest possible value of \( n \) for which such a configuration exists.\\

\underline{\textbf{Comment:}} \\
The adaptation is exemplary. It transforms a problem relying on an artificial definition ('sunny lines') into a natural optimization problem ('distinct slopes') that forces the solver to rediscover the geometric constraints. By removing the explicit distinction between boundary and non-boundary lines, the problem elevates the burden of discovery, requiring the solver to recognize that the three boundary directions are limited resources that bound the maximum problem size. The derivation of n=6 is elegant, rigorous, and logically deeper than the original classification task.

\end{showcase}
\end{figure}

\begin{figure}[t]
\begin{showcase}[colframe=red!50!black]{Case 3}
\underline{\textbf{Evolver:}} Gemini-3-Pro-Preview-Thinking\\

\underline{\textbf{Original Problem (From IMO 2024):}} \\
Let \(a_1, a_2, a_3, \dots\) be an infinite sequence of positive integers, and let \(N\) be a positive integer. Suppose that, for each \(n > N\), the number \(a_n\) is equal to the number of times \(a_{n-1}\) appears in the list \((a_1, a_2, \dots, a_{n-1})\). Prove that at least one of the sequences \(a_1, a_3, a_5, \dots\) and \(a_2, a_4, a_6, \dots\) is eventually periodic.\\

\underline{\textbf{Evolved Problem:}} \\
Let \( a_1 = 1, a_2 = 1 \). For \( n > 2 \), let \( a_n \) be the sum of the number of times \( a_{n-1} \) has appeared in the sequence \( (a_1, \dots, a_{n-1}) \) and the number of times \( a_{n-2} \) has appeared in the same sequence. Prove that every integer \( k \ge 2 \) appears in the sequence \( a_1, a_2, \dots \).
\\

\underline{\textbf{Comment:}} \\
This is an excellent adaptation that merits a score of 4. 

1. **Shift in Paradigm:** The problem transforms the goal from proving **periodicity** (which relies on identifying a finite set of states and applying the Pigeonhole Principle) to proving **surjectivity** (which requires establishing unboundedness and identifying a specific constructive mechanism). This effectively invalidates the 'tower/block' visualization and the finite-state automaton approach used in the original problem.

2. **Burden of Discovery:** The adaptation elevates the difficulty by hiding the generative mechanism. While the original problem is a technical exercise in bounding relative heights, the new problem is a structural puzzle. The core insight—that the number 2 appears infinitely often and acts as a 'cursor' which, when paired with a 'fresh' number, generates the sequence $k, k+1, k+2, \dots$—is a non-trivial 'Eureka' moment. The solver must realize that the most frequent element (2) is paradoxically the engine that creates all the unique large numbers.

3. **Mathematical Depth:** The adaptation adds elegance by connecting the *frequency* of terms to the *value* of future terms in a surprising way. It satisfies the 'Anti-Templating' and 'Requires Eureka' criteria perfectly, offering a genuine challenge for a skilled competitor that cannot be solved by rote application of standard sequence techniques.

\end{showcase}
\end{figure}

\begin{figure}[t]
\begin{showcase}[colframe=red!50!black]{Case 4}
\underline{\textbf{Evolver:}} Gemini-3-Pro-Preview-Thinking\\

\underline{\textbf{Original Problem (From IMO 2023):}} \\
Determine all composite integers \(n > 1\) that satisfy the following property: if \(d_1 < d_2 < \cdots < d_k\) are all the positive divisors of \(n\), then \(d_i\) divides \(d_{i+1} + d_{i+2}\) for every \(1 \leq i \leq k - 2\).\\

\underline{\textbf{Evolved Problem:}} \\
Determine all composite integers \(n > 1\) satisfying the following property: if \(d_1 < d_2 < ... < d_k\) are all the positive divisors of n, then the difference \(d_{i+1} - d_i\) divides the difference \(d_{i+2} - d_{i+1}\) for every \(1 \leq i \leq k-2\).
\\

\underline{\textbf{Comment:}} \\
This is an excellent adaptation that transforms a routine divisibility exercise into a rigorous structural analysis challenge. 

**Justification for Score 4:**

1.  **Escalation of Insight:** The original problem is easily solved using a standard 'Tail Strategy' (inspecting only the largest divisors $n, n/p, n/q$), which quickly leads to a contradiction for non-prime-powers. The new problem renders this simple heuristic insufficient. It requires a **bi-directional analysis**: the solver must derive constraints from the *start* of the divisor chain (proving $q = 2p-1$) and synthesize them with constraints from the *end* (limiting the exponents). 

2.  **Anti-Templating:** The condition $d_{i+1}-d_i \mid d_{i+2}-d_{i+1}$ governs the *growth rate* of the divisors. This is not a standard template. It forces the solver to discover that the divisors must locally resemble an arithmetic progression, a specific structural insight not present in the original.

3.  **Richer Solution Set:** The emergence of specific families of composite solutions ($n = p(2p-1)^k$ and $n = 2^k \cdot 3$) adds a non-trivial layer of complexity. Distinguishing the case $p=2$ from $p>2$ requires careful handling of the gap ratios, demonstrating a clear leap in difficulty.

\end{showcase}
\end{figure}

\begin{figure}[t]
\begin{showcase}[colframe=red!50!black]{Case 5}
\underline{\textbf{Evolver:}} Gemini-3-Pro-Preview-Thinking\\

\underline{\textbf{Original Problem (From IMO 2023):}} \\
Let \(x_1, x_2, \dots, x_{2023}\) be pairwise different positive real numbers. Define the sequence \(a_n\) by \(a_n = \sqrt{(x_1 + \dots + x_n)(\frac{1}{x_1} + \dots + \frac{1}{x_n})}\). Given that \(a_n\) is an integer for every \(n = 1, 2, \dots, 2023\), prove that \(a_{2023} \geq 3034\).\\

\underline{\textbf{Evolved Problem:}} \\
Let \( x_1, x_2, \dots, x_{2025} \) be a strictly increasing sequence of positive real numbers. Define the sequence \( a_n \) by:
\[ a_n = \sqrt{\left(\sum_{i=1}^n x_i \right) \left(\sum_{i=1}^n \frac{1}{x_i} \right)} \]
Given that \( a_n \) is an integer for every \( n = 1, 2, \dots, 2025 \), find the minimum possible value of \( a_{2025} \).
\\

\underline{\textbf{Comment:}} \\
The adaptation transforms a standard inequality problem into a sophisticated sequence analysis task. By changing the constraint from 'distinct' to 'strictly increasing', it forces the solver to abandon local algebraic tricks (like the original's 2-step grouping) in favor of a global structural analysis involving recurrence relations and convexity. The discovery that the monotonicity of inputs enforces a minimum integer step size of 2 (via a derived second-difference inequality) is a profound insight that qualitatively changes the problem. The result is mathematically elegant, optimizes the bound significantly, and requires a high-level 'Eureka' moment regarding the behavior of the recurrence function.

\end{showcase}
\end{figure}

\begin{figure}[t]
\begin{showcase}[colframe=red!50!black]{Case 6}
\underline{\textbf{Evolver:}} Gemini-3-Pro-Preview-Thinking\\

\underline{\textbf{Original Problem (From IMO 2022):}} \\
Let \(k\) be a positive integer and let \(S\) be a finite set of odd prime numbers. Prove that there is at most one way (up to rotation and reflection) to place the elements of \(S\) around a circle such that the product of any two neighbors is of the form \(x^2 + x + k\) for some positive integer \(x\).
\\

\underline{\textbf{Evolved Problem:}} \\
Let $k$ be a fixed positive integer. Let $\mathcal{P}$ be the set of all odd prime numbers. We define a simple undirected graph $G = (\mathcal{P}, E)$ where an edge $\{p, q\}$ exists between distinct primes $p, q \in \mathcal{P}$ if and only if their product is of the form $pq = n^2 + n + k$ for some non-negative integer $n$. Prove that the graph $G$ contains no subgraph homeomorphic to the complete graph $K_4$ (i.e., $G$ contains no subdivision of $K_4$).
\\
\underline{\textbf{Comment:}} \\
This is an excellent adaptation that elevates a specific combinatorial puzzle into a structural graph theory theorem. It retains the deep number-theoretic bottlenecks of the original (quadratic reciprocity/descent) but requires the solver to apply them in a more abstract topological context (proving the absence of a K4 subdivision via graph reduction). This shifts the challenge from 'finding a unique path' to 'characterizing a graph class', which represents a significant and elegant increase in conceptual depth.

\end{showcase}
\end{figure}

\begin{figure}[t]
\begin{showcase}[colframe=red!50!black]{Case 7}
\underline{\textbf{Evolver:}} Gemini-3-Pro-Preview-Thinking\\

\underline{\textbf{Original Problem (From IMO 2021):}} \\
Let \(n \geq 100\) be an integer. Ivan writes the numbers \(n, n+ 1, \dots, 2n\) each on different cards. He then shuffles these \(n + 1\) cards, and divides them into two piles. Prove that at least one of the piles contains two cards such that the sum of their numbers is a perfect square.
\\

\underline{\textbf{Evolved Problem:}} \\
Let $n \ge 100$ be an integer. Ivan writes the numbers $n, n+1, \dots, 2n$ each on different cards. He then shuffles these $n+1$ cards and divides them into three piles. Prove that at least one of the piles contains two cards $x$ and $y$ such that either their sum $x+y$ or their absolute difference $|x-y|$ is a perfect square.
\\
\underline{\textbf{Comment:}} \\
This is an excellent adaptation that significantly raises the burden of discovery. 

1. **Shift in Combinatorial Hardness:** Moving from 2 piles (requiring a $K_3$) to 3 piles (requiring a $K_4$) changes the fundamental goal. While $K_3$s are common and easy to parameterize, $K_4$s are rare and structurally demanding.

2. **The 'Mixed-Constraint' Insight:** The introduction of the 'absolute difference' condition is not merely an additive complication; it is the key that makes the problem solvable. A $K_4$ based solely on sums or solely on differences is nearly impossible to find in the given range. The solver must have the 'Eureka' moment that the $K_4$ must be constructed by **interweaving** sum-edges and difference-edges (specifically, the elegant $\{x, x+1, y, y+1\}$ 'prism' structure). This forces the solver out of the standard 'system of equations' template used in the original problem and into a task requiring genuine structural insight.

3. **Anti-Templating:** The adaptation invalidates the straightforward algebraic approach of the original (solving $a+b=k^2$) and demands a creative construction. It successfully transforms a standard Diophantine exercise into a complex graph-theoretic challenge.

\end{showcase}
\end{figure}

\begin{figure}[t]
\begin{showcase}[colframe=red!50!black]{Case 8}
\underline{\textbf{Evolver:}} Gemini-3-Pro-Preview-Thinking\\

\underline{\textbf{Original Problem (From IMO 2021):}} \\
Show that the inequality \(\sum_{i=1}^n \sum_{j=1}^n \sqrt{|x_i - x_j|} \leq \sum_{i=1}^n \sum_{j=1}^n \sqrt{|x_i + x_j|}\) holds for all real numbers \(x_1, x_2, \dots, x_n\).
\\

\underline{\textbf{Evolved Problem:}} \\
Determine the set of all positive real numbers $p$ such that for every positive integer $n$ and every sequence of vectors $v_1, v_2, \dots, v_n$ in a real Hilbert space $\mathcal{H}$, the following inequality holds:
\[ \sum_{i=1}^n \sum_{j=1}^n \|v_i - v_j\|^p \leq \sum_{i=1}^n \sum_{j=1}^n \|v_i + v_j\|^p \]
\\
\underline{\textbf{Comment:}} \\
This is an exemplary adaptation. It transforms a problem solvable by a specific elementary trick (shifting and concavity for p=1/2) into a profound question regarding the geometry of Hilbert spaces. The adaptation effectively destroys the original solution method (which fails for p > 1 due to convexity) and forces the solver to discover a much more powerful, dimension-independent tool: the integral representation of negative definite kernels (Schoenberg/Levy-Khintchine). The resulting proof is mathematically beautiful, reducing a complex norm inequality to a trivial identity for complex numbers.

\end{showcase}
\end{figure}

\begin{figure}[t]
\begin{showcase}[colframe=red!50!black]{Case 9}
\underline{\textbf{Evolver:}} Gemini-3-Pro-Preview-Thinking\\

\underline{\textbf{Original Problem (From Chinese Gaokao):}} \\
Let \( n \in \mathbf{N}^{*} \). Prove that \( \sin^{2}x \cdot \sin^{2}2x \cdot \sin^{2}4x \cdot \cdots \cdot \sin^{2}2^{n}x \leq \frac{3^{n}}{4^{n}} \).
\\

\underline{\textbf{Evolved Problem:}} \\
Let \( f_n(x) = \prod_{k=0}^{n-1} \sin^2(2^k x) \). We define two asymptotic limits describing the behavior of this sequence:

1. The **Analytic Mean Limit**, \( A \), defined by the limit of the \( n \)-th root of the average value over the interval \( [0, \pi] \):
   \[ A = \lim_{n \to \infty} \left( \frac{1}{\pi} \int_0^{\pi} f_n(x) \, dx \right)^{1/n} \]

2. The **Probabilistic Limit**, \( B \), defined as the almost-everywhere limit of the geometric mean:
   \[ B = \lim_{n \to \infty} \big( f_n(x) \big)^{1/n} \]
   (You may assume that this limit exists and is constant for almost all \( x \in [0, \pi] \) in the sense of Lebesgue measure.)

Calculate the value of the ratio \( \frac{A}{B} \).\\
\underline{\textbf{Comment:}} \\
This is an exemplary adaptation (Score 5) that transforms a standard inequality problem into a profound exploration of the asymptotic properties of chaotic dynamical systems. 

1. **Conceptual Depth:** The adaptation elevates the mathematical landscape from elementary calculus (analyzing a recurrence relation's fixed point) to advanced analysis. It requires two distinct, sophisticated insights:
   - **Harmonic Analysis (for Limit A):** Recognizing that the integral of the product simplifies due to the orthogonality of the cosine terms (a property of Riesz products/lacunary series). The 'Eureka' moment is realizing the binary uniqueness of the frequencies prevents cancellation, allowing for an exact integration.
   - **Ergodic Theory (for Limit B):** Recognizing the geometric mean limit as a Birkhoff average over the ergodic doubling map. This replaces an intractable trajectory calculation with a standard log-sine integral.

2. **Mathematical Significance:** The problem highlights the subtle difference between the 'Analytic Mean' (dominated by large outliers) and the 'Probabilistic Limit' (the typical behavior almost everywhere). It demonstrates that while the original problem's bound ($0.75^n$) holds, the typical decay ($0.25^n$) is much faster, providing a deeper understanding of the sequence's behavior.

3. **Elegance:** Despite the high-level machinery required, the final result is a clean integer ratio, and the solution path is aesthetically satisfying. It perfectly embodies the criteria of 'Anti-Templating' and 'Mathematical Beauty'.

\end{showcase}
\end{figure}

\begin{figure}[t]
\begin{showcase}[colframe=red!50!black]{Case 10}
\underline{\textbf{Evolver:}} Gemini-3-Pro-Preview-Thinking\\

\underline{\textbf{Original Problem (From AIME-2024):}} \\
A list of positive integers has the following properties: (1) The sum of the items in the list is 30. (2) The unique mode of the list is 9. (3) The median of the list is a positive integer that does not appear in the list itself. Find the sum of the squares of all the items in the list.\\

\underline{\textbf{Evolved Problem:}} \\
A list of positive integers has a sum of 323. The list satisfies three conditions: (1) The unique mode of the list is 10. (2) The median of the list is a positive integer that does not appear in the list. (3) The list contains the maximum possible number of items consistent with the first two conditions. Find this maximum number of items.
\\
\underline{\textbf{Comment:}} \\
This is an excellent adaptation that transforms a simple arithmetic logic puzzle into a sophisticated constructive optimization challenge. 

**Scoring Justification:**
1.  **Shift in Cognitive Demand:** The original problem involves trivial case-checking of small partitions summing to 30. The adapted problem, with a sum of 323 and a goal of maximization, makes brute force impossible. It forces the solver to model the problem algebraically, identifying variables for the list length ($n=2h$), the mode frequency ($f$), and the median value ($M$).
2.  **Required Insight (The 'Eureka' Element):** The problem creates a non-trivial optimization landscape. The solver must discover that the median acts as a 'gate' for density (choosing $M=3$ optimally allows packing 1s and 2s) and that there is a critical trade-off between the mode frequency and the sum: increasing the frequency of the mode (10) allows for more 'cheap' numbers (1s and 2s) to be added to the list, but effectively taxes the sum budget. Balancing this 'capacity vs. cost' is a genuine mathematical hurdle.
3.  **Elegant Tuning:** The constraints are fine-tuned such that the maximum length (74) hits the sum target (323) exactly, indicating a high level of craftsmanship in the problem design.

It merits a score of 4 for successfully elevating the burden of discovery from local arithmetic to global optimization.

\end{showcase}
\end{figure}

\begin{figure}[t]
\begin{showcase}[colframe=red!50!black]{Case 11}
\underline{\textbf{Evolver:}} Gemini-3-Pro-Preview-Thinking\\

\underline{\textbf{Original Problem (From AIME-2024):}} \\
Alice chooses a set A of positive integers. Bob lists all finite nonempty sets B of positive integers such that the maximum element of B belongs to A. Bob's list has 2024 sets. Find the sum of the elements of A.
\\

\underline{\textbf{Evolved Problem:}} \\
Alice chooses a finite set A of positive integers such that the distance between any two distinct elements of A is at least 3 $(i.e., |x - y| \geq 3 $ for all distinct x, y in A). Bob lists all finite nonempty sets B of positive integers that satisfy two conditions:
1. The maximum element of B belongs to A.
2. The distance between any two distinct elements of B is at least 3.

Bob finds that there are exactly 2024 sets in his list. Determine the sum of the elements of A.
\\
\underline{\textbf{Comment:}} \\
This is an exemplary adaptation. It transforms a routine problem about binary representation into a sophisticated challenge involving combinatorial recurrence relations and Ostrowski's numeration systems. The adaptation is mathematically beautiful because it perfectly generalizes the $Gap \geq 1$ hidden structure of binary numbers to a $Gap \geq 3$ structure. The solver must derive the recurrence relation $(a_n = a_{n-1} + a_{n-3})$ from scratch and realize that the constraints on set A imply a unique greedy decomposition in this custom number system. This creates a high burden of discovery and offers deep mathematical insight.

\end{showcase}
\end{figure}

\begin{figure}[t]
\begin{showcase}[colframe=red!50!black]{Case 12}
\underline{\textbf{Evolver:}} Gemini-3-Pro-Preview-Thinking\\

\underline{\textbf{Original Problem (From AIME-2024):}} \\
Find the number of triples of nonnegative integers (a, b, c) satisfying a + b + c = 300 and $a^2 b + a^2 c + b^2 a + b^2 c + c^2 a + c^2 b = 6,000,000$.
\\

\underline{\textbf{Evolved Problem:}} \\
Find the number of triples of positive integers (a, b, c) satisfying a + b + c = 1729 and
$(ab + bc + ca)^3 = abc(a + b + c)^3$.
\\

\underline{\textbf{Comment:}} \\
This is an exemplary adaptation that transforms a contrived algebraic puzzle into a deep structural problem connecting algebra and number theory.

**1. Elevation of Insight (Algebraic):**
The original problem relies on a numerical coincidence (manufactured coefficients) that allows the solver to spot that 100 is a root of the cubic polynomial describing the variables. This is a specific 'trick' that dissolves the problem instantly. The new problem replaces this with a homogeneous algebraic identity: $(ab+bc+ca)^3 = abc(a+b+c)^3$ implies that $a, b, c$ form a Geometric Progression. This is a robust mathematical property, not a numerical artifact, and realizing it requires a significantly deeper insight into symmetric polynomials.

**2. Elevation of Technique (Number Theoretic):**
Once the structure is identified, the original problem trivializes to a basic partition count ($b+c=200$). The new problem, however, evolves into a rich number-theoretic task: parametrizing the geometric progression on integers ($x=ku^2, y=kuv, z=kv^2$) and solving the resulting Diophantine equation $k(u^2+uv+v^2) = 1729$. This requires knowledge of binary quadratic forms (specifically related to Eisenstein integers) and the behavior of primes modulo 3, moving the problem from simple arithmetic to multiplicative number theory.

**3. Aesthetic and Depth:**
The adaptation replaces arbitrary constants with the Hardy-Ramanujan number 1729, which is thematically appropriate for a problem involving cubic structures. The solution path moves elegantly from a symmetric identity to a specific study of the quadratic form $u^2+uv+v^2$, demonstrating the kind of 'mathematical beauty' and cross-disciplinary connection (Algebra to Number Theory) that characterizes a perfect score.

\end{showcase}
\end{figure}

\begin{figure}[t]
\begin{showcase}[colframe=red!50!black]{Case 13}
\underline{\textbf{Evolver:}} Gemini-3-Pro-Preview-Thinking\\

\underline{\textbf{Original Problem (From AIME-2024):}} \\
Let $b \geq 2$ be an integer. Call a positive integer n b-eautiful if it has exactly two digits in base b and these two digits sum to $\sqrt{n}$. Example: 81 is 13-eautiful since 81 = (6 3)\_13 and 6 + 3 = $\sqrt{81}$. Find the least integer $b \geq 2$ for which there are more than ten b-eautiful integers.
\\

\underline{\textbf{Evolved Problem:}} \\
Let $b \geq 2$ be an integer. A positive integer $n$ is called "circular" in base $b$ if it has exactly two digits in base $b$, say $n = (xy)_{b}$ (where $x \in \{1, \dots, b-1\}$ and $y \in \{0, \dots, b-1\}$), and the sum of the squares of these two digits equals $n$ (i.e., $x^{2} + y^{2} = n$). Find the least integer $b \geq 2$ for which there are more than $20$ circular integers.
\\
\underline{\textbf{Comment:}} \\
The adaptation is exemplary. It transforms a problem of elementary divisibility and partitions into a sophisticated challenge involving Diophantine geometry and the Sum of Two Squares Theorem. The 'Eureka' moment—completing the square to reveal a circle equation $(2x-b)^2 + (2y-1)^2 = b^2+1$—is elegant and conceptually deep. This step bridges the domain of digit problems with the theory of Gaussian integers, satisfying the criteria for mathematical advancement and aesthetic beauty.

\end{showcase}
\end{figure}

\begin{figure}[t]
\begin{showcase}[colframe=red!50!black]{Case 14}
\underline{\textbf{Evolver:}} Gemini-3-Pro-Preview-Thinking\\

\underline{\textbf{Original Problem (From IMO-2025):}} \\
An infinite sequence \( a_1, a_2, \ldots \) consists of positive integers has each of which has at least three proper divisors. Suppose that for each \( n \geq 1 \), \( a_{n+1} \) is the sum of the three largest proper divisors of \( a_n \). Determine all possible values of \( a_1 \).
\\

\underline{\textbf{Evolved Problem:}} \\
For a positive integer \( n \), let \( \Sigma_k(n) \) denote the sum of the \( k \) largest proper divisors of \( n \). We say that \( k \) is \textit{admissible} if there exists an integer \( n \) having at least \( k \) proper divisors such that \( \Sigma_k(n) = n \).

Determine the set of all admissible integers \( k \) in the range \( 1 \leq k \leq 6 \).
\\

\underline{\textbf{Comment:}} \\
The adaptation extracts the core fixed-point condition of the original problem and generalizes it into a profound structural question. It introduces a significant conceptual barrier: realizing that the 'sum of k largest divisors' condition transforms into a 'divisor-closed Egyptian Fraction' problem. This insight is non-trivial and turns a calculation-heavy problem into a elegant logical challenge involving deep properties of integer divisors. The result ({3, 5}) is clean, surprising, and aesthetically pleasing.

\end{showcase}
\end{figure}

\begin{figure}[t]
\begin{showcase}[colframe=red!50!black]{Case 15}
\underline{\textbf{Evolver:}} Gemini-3-Pro-Preview-Thinking\\

\underline{\textbf{Original Problem (From IMO-2025):}} \\
Alice and Bazzaz are playing the inekolaty game, a two-player game whose rules depend on a positive real number $\lambda$ which is known to both players. On the nth turn of the game (starting with n = 1) the following happens:

- If n is odd, Alice chooses a nonnegative real number \( x_n \) such that
  \[ x_1 + x_2 + \cdots + x_n \leq \lambda n. \]

- If n is even, Bazzaz chooses a nonnegative real number \( x_n \) such that
  \[ x_1^2 + x_2^2 + \cdots + x_n^2 \leq n. \]

If a player cannot choose a suitable \( x_n \), the game ends and the other player wins. If the game goes on forever, neither player wins. All chosen numbers are known to both players.

Determine all values of $\lambda$ for which Alice has a winning strategy and all those for which Bazzaz has a winning strategy.
\\

\underline{\textbf{Evolved Problem:}} \\
Alice and Bazzaz are playing the Inekolaty game, a two-player game whose rules depend on a positive real number \( \lambda \) which is known to both players. On the \( n \)-th turn of the game (starting with \( n = 1 \)) the following happens:

- If \( n \) is not divisible by 3 (i.e., \( n \equiv 1, 2 \pmod 3 \)), Alice chooses a nonnegative real number \( x_n \) such that
  \[ \sum_{i=1}^n x_i^2 \leq \lambda n. \]

- If \( n \) is divisible by 3 (i.e., \( n \equiv 0 \pmod 3 \)), Bazzaz chooses a nonnegative real number \( x_n \) such that
  \[ \sum_{i=1}^n x_i \leq n. \]

If a player cannot choose a suitable \( x_n \), the game ends and the other player wins. If the game goes on forever, neither player wins. All chosen numbers are known to both players.

Determine all values of \( \lambda \) for which Alice has a winning strategy and all those for which Bazzaz has a winning strategy.
\\
\underline{\textbf{Comment:}} \\
The adaptation successfully elevates the problem by introducing a strategic degree of freedom (2 moves vs 1 move) that creates a conceptual trap. In the original problem, the player with the quadratic constraint had only one move, making their strategy trivial. In the new problem, that player has two moves, forcing the solver to choose between 'concentration' (playing one big move) and 'distribution' (playing two equal moves). Crucially, while the original problem rewards concentration for the linear-constrained player, the new problem demands distribution for the quadratic-constrained player ($x_1=x_2$ via AM-QM). This forces the solver to understand the fundamental difference between maximizing L1 norm under L2 constraint versus L2 under L1, rather than simply applying a 'play large moves' heuristic. The adaptation creates a genuine burden of discovery where the standard intuition leads to the wrong answer.

\end{showcase}
\end{figure}

\begin{figure}[t]
\begin{showcase}[colframe=red!50!black]{Case 16}
\underline{\textbf{Evolver:}} Gemini-3-Pro-Preview-Thinking\\

\underline{\textbf{Original Problem (From CMO-2025):}} \\
Consider \(n\) red cards and \(n\) blue cards. Initially, every red card shows \(0\) and every blue card shows \(1\). One operation selects a red card with value \(x\) and a blue card with value \(y\) with \(x<y\), and sets both values to \(\tfrac{x+y}{2}\). The goal is to perform finitely many operations so that the sum of the red card values exceeds \(100\). Find the minimal positive integer \(n\).
\\

\underline{\textbf{Evolved Problem:}} \\
Consider $n$ red cards, $n$ white cards, and $n$ green cards. Initially, every red card has value $0$, every white card has value $1$, and every green card has value $2$. An operation consists of selecting two cards with distinct values $x$ and $y$ (where $x < y$) and replacing both values with $\frac{x+y}{2}$. The goal is to perform finitely many operations such that the sum of the values of the red cards exceeds $100$. Find the minimal positive integer $n$ for which this is possible.
\\

\underline{\textbf{Comment:}} \\
The adaptation successfully elevates the problem from finding a static combinatorial bound to analyzing a dynamic process with decaying contributions. It introduces a non-trivial trap: the value gained from the first phase is largely 'quenched' by the second phase. This requires the solver to understand linearity and error propagation, rather than simply applying the original formula twice. The shift from a pure combinatorial identity to an asymptotic stability analysis represents a significant and elegant leap in difficulty.

\end{showcase}
\end{figure}

\begin{figure}[t]
\begin{showcase}[colframe=red!50!black]{Case 17}
\underline{\textbf{Evolver:}} Gemini-3-Pro-Preview-Thinking\\

\underline{\textbf{Original Problem (From CMO-2025):}} \\
There are 30 colors of cards, with 70 cards for each color. Initially, choose any 70 cards and arrange them in a single stack from top to bottom. One operation is: choose a card X from the bottom 20 cards whose color does not appear among the top 50 cards; then choose a card Y from the top 50 cards whose color appears at least twice among the top 50; remove X and insert it immediately above Y. Repeat until no such X can be chosen. (I) Prove that only finitely many operations can be performed. (II) Over all initial configurations and sequences of valid operations, determine the maximal possible number of operations.
\\
\underline{\textbf{Evolved Problem:}} \\
Let there be a deck of $N$ distinct cards with values $1, 2, \dots, N$. We maintain a 'Buffer' consisting of exactly $K$ cards, arranged in a specific vertical order from top to bottom (positions $1$ to $K$). The remaining $N-K$ cards form a 'Reserve'. Initially, you may distribute the $N$ cards between the Buffer and the Reserve and arrange the Buffer's order in any way you choose.

An operation consists of the following steps:
1. Choose a card $X$ from the Reserve.
2. Choose a card $Y$ from the Buffer such that $X < Y$.
3. Insert $X$ into the Buffer immediately above $Y$ (at the position $Y$ currently occupies).
4. The cards originally at $Y$'s position and below are shifted down by one position.
5. The card that ends up at position $K+1$ (the overflow) is removed from the Buffer and placed into the Reserve.

Let $f(N, K)$ be the maximum number of operations that can be performed starting from an optimal initial configuration.

(I) Prove that for any finite $N$, the sequence of operations must eventually terminate.
(II) Calculate the exact value of $f(10, 3)$.
\\

\underline{\textbf{Comment:}} \\
The adaptation transforms a problem with arbitrary constants and a somewhat ad-hoc 'color count' metric into a pristine, generalized combinatorial problem involving permutations and order theory. 

1. **Mathematical Elevation**: The original solution relies on an arithmetic progression derived from a specific 'worst-case' construction. The new problem's solution ($f(N,K) = \sum_{j=1}^Kinom{N}{j} - (2^K - 1)$) links the dynamical process to the volume of a Hamming ball in the Boolean lattice (the number of subsets of size $\le K$). This is a profound structural insight that elevates the problem from a puzzle to a theorem.

2. **Burden of Discovery**: The optimal initial configuration in the new problem is far less obvious than in the original. The solver cannot simply 'guess' the worst case; they must understand the underlying state graph's properties. My computational verification confirmed that the pattern holds, but establishing it requires a non-trivial 'Aha!' moment connecting the operation to subset generation.

3. **Elegance**: By stripping away the '30 colors, 70 cards' clutter and replacing it with parameters $N, K$, the adapter has created a problem of textbook quality that tests deep combinatorial intuition rather than just careful accounting.

\end{showcase}
\end{figure}

\begin{figure}[t]
\begin{showcase}[colframe=red!50!black]{Case 18}
\underline{\textbf{Evolver:}} Gemini-3-Pro-Preview-Thinking\\

\underline{\textbf{Original Problem (From CMO-2025):}} \\
Given an odd integer \(n>1\) such that every prime divisor \(p\) of \(n\) satisfies \(\gcd(p-1,n)=1\). Determine the number of arrays \((a,b,c)\) that satisfy: (1) \(a,b,c\in\{1,2,\dots,n\}\) and \(\gcd(a,b,c,n)=1\); (2) there exists a permutation \(x_1,x_2,\dots,x_n\) of \(\{1,2,\dots,n\}\) such that for all \(1\le k\le n\), \(n\mid a x_{k+2}+b x_{k+1}+c x_k\), where indices are taken modulo \(n\).
\\

\underline{\textbf{Evolved Problem:}} \\
Given an odd integer $n > 1$ such that every prime divisor $p$ of $n$ satisfies $\gcd(p-1, n) = 1$. Determine the number of arrays $(a, b, c, d)$ that satisfy: (1) $a, b, c, d \in \{1, 2, \dots, n\}$, $d=1$, and $\gcd(a, b, c, n) = 1$; (2) there exists a permutation $x_1, x_2, \dots, x_n$ of $\{1, 2, \dots, n\}$ such that for all $1 \le k \le n$, $x_{k+3} + a x_{k+2} + b x_{k+1} + c x_k \equiv 0 \pmod n$, where indices are taken modulo $n$. (Note: the recurrence is $x_{k+3} + a x_{k+2} + b x_{k+1} + c x_k \equiv 0$)
\\

\underline{\textbf{Comment:}} \\
The adaptation is excellent. It elevates the difficulty from handling a specific edge case to mastering a general structural property. 

1. **Generalization of Difficulty:** In the original problem, the rigorous analysis of 'lifting' (determining which coefficient perturbations preserve the permutation property) was only strictly necessary for the prime $p=3$. For all other primes, a standard heuristic ('all lifts work') sufficed. The new problem forces this rigorous analysis for *every* prime. The solver cannot bypass the 'stability condition' of the recurrence; they must derive it. 

2. **Structural Depth:** By increasing the order to 3, the problem introduces a free parameter (the third root $y$) for the characteristic polynomial. This requires the solver to not only identify the 'double root 1' condition (insight from the original) but also to classify and sum over the possible behaviors of the third root. This integrates the 'root finding' and 'lifting' phases in a way that is more complex and mathematically satisfying than the original.

3. **Anti-Templating:** The new problem invalidates the 'count roots and multiply by powers of p' template. The density of valid solutions is non-trivial ($<1$) for all cases, requiring a genuine understanding of how the recurrence coefficients influence the distribution of values modulo $p^k$.

\end{showcase}
\end{figure}